\documentclass{article}
\usepackage[numbers]{natbib}
\usepackage{arxiv}
\usepackage{times}  % DO NOT CHANGE THIS
\usepackage{helvet}  % DO NOT CHANGE THIS
\usepackage{courier}  % DO NOT CHANGE THIS
\usepackage[hyphens]{url}  % DO NOT CHANGE THIS
\usepackage{graphicx} % DO NOT CHANGE THIS
\usepackage{hyperref}       % hyperlinks
\usepackage{url}            % simple URL typesetting
\usepackage{booktabs}       % professional-quality tables
\usepackage{amsfonts}       % blackboard math symbols
\usepackage{nicefrac}       % compact symbols for 1/2, etc.
\usepackage{microtype}      % 
\usepackage{xcolor}         % colors
\usepackage{graphicx}
\usepackage{xspace}
\usepackage{mathtools}
\usepackage{amssymb,mathrsfs,amsmath}
\usepackage{pifont}
\usepackage{amsthm}
\usepackage{color}
\usepackage{amsfonts}
\usepackage{subcaption}
\usepackage{multirow}
\newtheorem{theorem}{Theorem}
\usepackage[ruled, vlined, linesnumbered]{algorithm2e}

\usepackage{cleveref} 
\usepackage{letltxmacro}
\LetLtxMacro{\originaleqref}{\eqref}
\renewcommand{\eqref}{\text{Eq.}~\originaleqref}
\crefname{figure}{Fig.}{Figs.}
\Crefname{figure}{Figure}{Figures}

\crefname{table}{Table}{Tables}
\Crefname{table}{Table}{Tables}

\crefname{section}{Sec.}{Secs.}
\Crefname{section}{Section}{Sections}

\usepackage{newfloat}
\usepackage{listings}

\usepackage{authblk}
\newcommand\nnfootnote[1]{%
  \begin{NoHyper}
  \renewcommand\thefootnote{}\footnote{#1}%
  \addtocounter{footnote}{-1}%
  \end{NoHyper}
}
\author[]{\textbf{Fan-Ming Luo}}
\author[]{\textbf{Xingchen Cao}}
\author[]{\textbf{Rong-Jun Qin}}
\author[]{\textbf{Yang Yu}\textsuperscript{\dag}}
\affil[]{National Key Laboratory for Novel Software Technology, Nanjing University, China\\School of Artificial Intelligence, Nanjing University, China\\Polixir.ai\\\texttt{\{luofm, caoxc, qinrj, yuy\}@lamda.nju.edu.cn}}
\date{}

\title{Transferable Reward Learning by Dynamics-Agnostic Discriminator Ensemble}

\begin{document}

% Title
\maketitle

\begin{abstract}
Recovering reward function from expert demonstrations is a fundamental problem in reinforcement learning. The recovered reward function captures the motivation of the expert. Agents can imitate experts by following these reward functions in their environment, which is known as \emph{apprentice learning}. However, the agents may face environments different from the demonstrations, and therefore, desire transferable reward functions. Classical reward learning methods such as inverse reinforcement learning (IRL) or, equivalently, adversarial imitation learning (AIL), recover reward functions coupled with training dynamics, which are hard to be transferable. Previous dynamics-agnostic reward learning methods rely on assumptions such as that the reward function has to be state-only, restricting their applicability. In this work, we present a dynamics-agnostic discriminator-ensemble reward learning method (DARL) within the AIL framework, capable of learning both state-action and state-only reward functions. DARL achieves this by decoupling the reward function from training dynamics, employing a dynamics-agnostic discriminator on a latent space derived from the original state-action space. This latent space is optimized to minimize information on the dynamics. We moreover discover the policy-dependency issue of the AIL framework that reduces the transferability. DARL represents the reward function as an ensemble of discriminators during training to eliminate policy dependencies. Empirical studies on MuJoCo tasks with changed dynamics show that DARL better recovers the reward function and results in better imitation performance in transferred environments, handling both state-only and state-action reward scenarios.
\end{abstract}
\nnfootnote{\dag: Yang Yu is the corresponding author.}
\newcommand{\ours}{DARL\xspace}

\section{Introduction}
Rewards are at the core of Markov decision processes (MDPs)~\cite{sutton1998rl}, representing the underlying incentives for optimal behavior~\cite{kim2021identify}. Inverse reinforcement learning (IRL)~\cite{ng2000irl,ng2004apprentice,russell1998learning} excels in inferring reward functions from expert demonstrations, eliminating the need for manually crafted rewards~\cite{lee2019agile,kumar2021rma}. This technique has numerous applications~\cite{kim2021identify}, such as deducing the incentives of real-world decision-makers~\cite{fahad2018pedestrian,niv2009reinforcement,ashwood2022dynamic} and enabling advanced counterfactual reasoning~\cite{rust1994structural,kalouptsidi2021identification,christensen2019counterfactual}. By recovering reward functions from the demonstrations of professionals such as financiers or clinicians, we can better understand the dynamics of financial markets and healthcare systems. Moreover, these reward functions can facilitate the prediction of expert behaviors in new scenarios, such as predicting the treatment decisions of a physician for a new patient. These applications require reward functions that are not only generalizable but also encapsulating the core motivations of the experts. However, existing IRL methods often struggle to derive reward functions that meet these criteria in diverse contexts~\cite{kim2021identify}.

Various methods have been developed to recover true or transferable reward functions that are robust against transition disturbances~\cite{ni2020firl,fu2017airl,geng2020pqr}.
These methods often rely on restrictive assumptions about the true reward function or the problem settings. For instance, they may assume that partial knowledge of the reward function is acquirable~\cite{geng2020pqr}, that the reward is exclusively state-dependent~\cite{ni2020firl,fu2017airl}, or that the environment dynamics can be freely adjusted~\cite{amin2016resolving}. Such assumptions may not always hold true, potentially limiting the applicability of these methods. In this work, we aim to learn a transferable reward function without imposing additional assumptions. Our approach is based on the established framework of adversarial imitation learning~\cite{ho2016gail,kostrikov2020valuedice}, offering a more universally applicable solution.

Adversarial imitation learning (AIL), a special type of IRL method~\cite{ni2020firl}, has gained significant attention for its simplicity and efficiency. AIL uses a discriminator to align the agent's occupancy measure with that of the expert by alternating between learning the discriminator and updating the policy. The discriminator aims to distinguish the trajectories generated by the policy from the expert demonstrations. The policy is then optimized to confuse the discriminator by maximizing the rewards derived from it. The process converges once the discriminator can no longer differentiate between the two data sources. Despite its effectiveness, AIL faces a critical limitation: the reward signal inferred from the discriminator lacks transferability and cannot be applied to downstream tasks~\cite{ni2020firl}, such as relearning policies from scratch in the original or an alternative environment.

In this work, we analyze the factors contributing to the limited generalizability of the reward functions obtained via AIL. The issues are primarily twofold: (1) The learned reward is dynamics-dependent~\cite{fu2017airl} because of reward ambiguity caused by reward shaping~\cite{ng1999shaping}, which transforms the reward while preserving policy optimality. Within the IRL framework, distinguishing whether a reward function has been altered by shaping remains a significant challenge, hindering the development of a reward function free from such modifications. Reward shaping is intrinsically linked to the original environment dynamics and fails to maintain policy optimality across differing dynamics~\cite{fu2017airl}, thereby constraining the transferability of the shaped reward. (2) The learned reward is also policy-dependent, arising from the iterative learning nature of AIL. The discriminator can only provide correct rewards for a narrow spectrum of policies. For instance, the discriminator at the convergence could fail to distinguish the policies learned at the early stages due to forgetting, but it retains the capability to discern and appropriately reward policies learned at later training stages, just prior to convergence.

Regarding the two problems of AIL, we propose \textit{Dynamics-Agnostic Discriminator-Ensemble Reward Learning} (\ours) to eliminate the dependence on both policies and dynamics. \ours produces a state-action reward that is robust and transferable. To minimize the dynamics dependency, \ours encodes the state and action into a latent space using a state encoder and an action encoder. The discriminator is optimized to distinguish generated data from expert data within this latent space. The encoders are designed to minimize mutual information between their outputs and environment dynamics, preventing the discriminator from inferring rewards based on dynamics information, particularly the next state. To address policy dependency, \ours represents the reward as an ensemble of the historical discriminators trained during AIL.  We theoretically substantiate that the discriminator ensemble can converge to an optimal classifier, capable of distinguishing all past policies from the expert policy. We evaluate \ours across five MuJoCo environments~\cite{todorov2012mujoco} with four types of dynamics transfers, including dynamics parameter perturbations and limb impairments. Compared to state-of-the-art AIL and IRL methods, \ours demonstrates superior ability to learn a state-action reward function that closely aligns with the true environment reward. This alignment results in higher returns across diverse tasks, in both state-only and state-action scenarios. Further ablation studies confirm the effectiveness of each component of \ours. In conclusion, our contributions are fourfold:
\begin{enumerate}
\item We propose a mutual information minimization approach to eliminate the dependency of the reward function on environment dynamics.
\item We disclose the limitations of existing AIL methods that learn policy-dependent discriminators, which hinder the transferability of the reward function to new environments. 
\item We introduce an ensemble of discriminators approach, designed to eliminate the dependency on policies.
\item We empirically validate that DARL reconstructs a more accurate reward function across different tasks, leading to higher policy final returns in environments with dynamics transfer.
\end{enumerate}

\section{Preliminaries}
\label{sec_preliminaries}
\noindent{\textbf{Reinforcement learning.}} An RL task $\mathcal{M}$ is often formalized as a Markov decision process (MDP)~\cite{sutton1998rl}, described by a tuple $\langle\mathcal{S}, \mathcal{A}, p, r, \gamma, \rho_0\rangle$, 
where $\mathcal{S}$ is the state space, 
$\mathcal{A}$ is the action space, $p$ is the transition distribution that maps $(s_t,a_t)$ to $s_{t+1}$ with probability $p(s_{t+1}| s_t,a_t)$, 
$r(s_t, a_t): \mathcal{S}\times\mathcal{A}\rightarrow \mathbb{R}$ is the reward function,
$\gamma\in(0,1)$ is the discount factor, 
and
$\rho_0(s_0)$ is the initial state distribution.
At each time step $t$, the RL agent observes a state $s_t$ and chooses an action $a_t$ following a policy $\pi(a_t| s_t)$.
Then the agent will observe a new state $s_{t+1}$ following $p(s_{t+1}| s_t,a_t)$, and get an immediate reward $r(s_t, a_t)$. 
$U_t^{r,p} = \sum_{i=0}^\infty\gamma^ir(s_{t+i},a_{t+i})$ is the discounted accumulated reward, a.k.a. return. 
We denote the transition function as $\mathcal{T}(s_t,a_t):\mathcal{S}\times\mathcal{A}\rightarrow\mathcal{S}$ in the MDPs with deterministic transitions.
The objective of RL is to find a policy $\pi$ that maximizes the expectation of return, i.e., 
\begin{equation}
\label{eq_rl_target}
\mathop{\max}_{\pi} ~~  J^{r,p}(\pi)= ~ \mathbb{E}_{ \rho_0(s_0),\pi(a_t| s_t)} \left[U_0^{r,p}\right].
\end{equation}

\noindent{\textbf{Adversarial imitation learning (AIL).}}
AIL is designed to derive an expert policy $\pi^E$ from an expert dataset $\mathcal{B}^E$ through adversarial processes~\cite{ho2016gail,fu2017airl,peng2019vdb}. A notable variant, Generative Adversarial Imitation Learning (GAIL), effectively recovers policies using minimal expert data~\cite{ho2016gail}. The GAIL framework consists of two components: a policy model $\pi(a| s)$ and a discriminator $D(s,a): \mathcal{S}\times\mathcal{A}\rightarrow [0,1]$. The target of the discriminator role is to differentiate between expert data and data sampled from the policy. Concurrently, the policy is trained to mimic the expert, aiming to fool the discriminator so it cannot distinguish between expert and generated data. This adversarial procedure involves the discriminator minimizing a cross-entropy loss function, while the policy maximizes its reward signal, defined as $\hat{r}(s,a) = -\log(1 - D(s,a))$.
By denoting $\mathcal{B}^\pi$ the data sampled by the policy $\pi$, the GAIL objective function is formalized as:
\begin{equation}
    \label{eq_gail_loss}
    \begin{aligned}
    \mathcal{L}(D, \pi) &= \lambda \mathcal{H}(\pi)-\mathbb{E}_{s,a\sim\mathcal{B}^E}[\log (D(s,a))] \\
    &~~~~~~~~~~~~~~~~-\mathbb{E}_{s,a\sim\mathcal{B}^\pi}[\log(1-D(s,a))] ,
    \end{aligned}
\end{equation}
where $\mathcal{H}(\pi)$ represents the entropy of the policy, and $\lambda$ is a regularization factor. The optimization objectives for $\pi$ and $D$ are to maximize and minimize $\mathcal{L}(D, \pi)$, respectively.

\section{Related Work}
 \noindent{\textbf{Inverse reinforcement learning (IRL).}} IRL recovers reward functions from expert behaviors~\cite{ng2000irl,ng2004apprentice}. The IRL process typically alternates between optimizing a policy that maximizes cumulative rewards and updating the reward function using both expert demonstrations and policy-generated data. Distinctions among IRL methods primarily arise in the reward update phase. In apprenticeship learning, for instance, the reward function is optimized to maximize the performance margin between the expert data and the learned policy~\cite{ng2004apprentice}. Alternatively, the MaxEnt IRL framework addresses reward updates by formulating them as a maximum likelihood estimation problem, incorporating a maximum entropy principle to ensure a stochastic representation of policy behavior~\cite{ziebart2008maxent}. Subsequently, Finn et al.~\cite{finn2016gcl} expanded on MaxEnt IRL with Guided Cost Learning (GCL), using neural networks to estimate reward functions and adapting the approach to complex, high-dimensional environments. More recently, Adversarial Imitation Learning (AIL) has emerged as an offshoot of IRL~\cite{ho2016gail,finn2016connection,fu2017airl,kostrikov2020valuedice,sun2021softdice,kostrikov2019dac}, learning rewards through a fully adversarial process. A discriminator is trained to distinguish between policy-generated and expert data, effectively using the discriminator's outputs as proxy rewards in the AIL frameworks. AIL has demonstrated high efficiency in replicating expert policies with few expert demonstrations~\cite{ho2016gail}. However, Ni et al.~\cite{ni2020firl} noted that such discriminators yield non-stationary rewards, which are not suitable as reward functions for training new policies from scratch. Our work adopts the AIL paradigm but makes a significant advancement: our model learns stationary rewards that are consistent with the true environment rewards. Moreover, the rewards are robust enough to facilitate policy training from scratch, even in environments with dynamics transfer. 

\noindent{\textbf{Robust reward learning.}} Our work aims to learn a robust reward function that is transferable across environments with varied dynamics. Traditionally, reward recovery has been viewed as an ill-posed problem: given a series of expert demonstrations, numerous potential rewards could rationalize the observed behavior. It has been shown that an optimal policy with respect to a reward function $r(s_t, a_t, s_{t+1})$ remains optimal for a modified reward $\tilde{r}_\Phi(s_t, a_t, s_{t+1}) = r(s_t, a_t, s_{t+1}) + \gamma \Phi(s_{t+1}) - \Phi(s_t)$, where $\Phi: \mathcal{S} \rightarrow \mathbb{R}$ represents an arbitrary state-dependent function~\cite{ng1999shaping}. This implies the existence of multiple reward functions that can equally explain expert behavior, leading to a multiplicity of solutions in IRL. Recent advancements, however, suggest that constraining the problem space can facilitate the identification of the `true' reward~\cite{kim2021identify,cao20221identify,skalse2022part_identify}. Amin and Singh~\cite{amin2016resolving} propose that modifying the transition dynamics and acquiring the expert demonstrations of the modified dynamics can lead to the recovery of the true environment reward, though such manipulations are often impractical. Furthermore, Jacq et al.~\cite{jacq2019learnfromlearner} suggest that true reward recovery is also possible by analyzing the trajectories of an agent being trained via maximum-entropy reinforcement learning techniques. Several approaches assume specific characteristics of the reward function: Deep PQR~\cite{geng2020pqr} assumes known rewards for a constant `anchor' action across all states, while Adversarial IRL (AIRL)~\cite{fu2017airl} and $f$-IRL~\cite{ni2020firl} hypothesize that the true reward depends solely on states. \ours differentiates itself from these approaches by adhering to a classic IRL framework without presupposing the form of the reward function or making assumptions about the expert data.

\noindent{\textbf{Transfer imitation learning.}}
Transfer IL and \ours both deal with varying learning conditions, but focus on different aspects. \textit{Cross-domain imitation learning} focuses on changes in state or action spaces, such as in robot control where additional joints enlarge these spaces~\cite{raychaudhuri2021cross_domain}, or in tasks with image viewpoint changes~\cite{stadie2017third_person}. A common approach to these problems is aligning data between domains~\cite{raychaudhuri2021cross_domain,kim2020domain_adaptive}, followed by IL with the expert data aligned to the target domain. Another approach maps data from different domains into a shared latent space and performs IL in this latent space~\cite{franzmeyer2022cross_domain,stadie2017third_person,cetin2021domain_robust,yin2022cross_domain}. 
Fickinge et al.~\cite{fickinger2022cross_domain} also use Gromov-Wasserstein distances~\cite{moli2011gromov,melis2018gromov} to directly facilitate cross-domain IL. 

The \textit{dynamics mismatch} between the expert and the learning agent~\cite{chae2022robust_imitation,gangwani2020_state_only} is another core issue in Transfer IL, being more relevant to our setting. Previous studies typically learn the optimal state transition function without action input, then use an inverse dynamics model to predict the optimal actions under the target dynamics based on the learned optimal state transitions~\cite{torabi2018bco,torabi2018gaifo,liu2020state_alignment}. Recent research also introduces action perturbations during policy learning to increase robustness to dynamics changes~\cite{viano2022robust_learning}. However, the focal point of transfer IL lies in policy adaptation, while our work primarily aims to recover a reward function, which represents a significant departure. The reward function is a more fundamental concept within within MDPs. It not only facilitates policy learning but also provides insights into expert behavior~\cite{ashwood2022irl_animal}, supports counterfactual reasoning~\cite{kalouptsidi2021identification}, etc.

\noindent{\textbf{Mutual information.}} 
\ours adopts the concept of mutual information (MI) minimization~\cite{cheng2020vclub}, a metric that quantifies the correlation between two random variables from an information theory perspective. MI has been extensively applied in unsupervised learning domains, such as variational autoencoders~\cite{kingma2014vae,higgins2017betavae,chen2016infogan}. In RL, MI has been leveraged for unsupervised skill discovery~\cite{eysenbach2019diayn} and exploration~\cite{kim2019emi}, and in imitation learning to mimic multi-modal expert data~\cite{li2017infogail}. In variational AIL (VAIL)~\cite{peng2019vdb}, MI between the input and the output of an intermediate layer of the discriminator is constrained, which stabilizes the adversarial training process by limiting information flow. \ours diverges from VAIL as \ours minimizes the MI between the intermediate output of the discriminator and the next state, rather than between the input and the discriminator.

\section{Issues with the Reward Learned by AIL}
\label{sec_matters}

In this section, we explore the limitations of the reward function generated by the discriminator in AIL. We argue that this function does not represent a true or transferrable reward, hindering the performance of downstream tasks. Specifically, it is inadequate for relearning an optimal policy, particularly in environments with dynamics that differ from those of the training environment. Our analysis identifies two main factors contributing to this shortcoming: the learned reward function is not only \textit{dynamics-dependent} but also \textit{policy-dependent}. 

%%%%%%%%%%%%%%%%%%%%%%%%%%%%%%%% Dynamics Dependency %%%%%%%%%%%%%%%%%%%%%%%%%%%%%%%%
\subsection{Dynamics Dependency}
The reward learned in AIL is policy-dependent, which exists as a common issue of IRL and has already been previously identified in AIRL~\cite{fu2017airl}. This issue results from the inherent `reward ambiguity' in IRL: multiple reward functions can rationalize the same expert policy and demonstrations, making the problem intrinsically ill-conditioned. It has been proved that an optimal policy w.r.t. a reward $r(s_t,a_t,s_{t+1})$ remains optimal for any reward function reshaped as
\begin{equation}
    \label{eq_reward_shaping}
    \tilde{r}(s_t,a_t,s_{t+1})=r(s_t,a_t,s_{t+1})+\gamma\Phi(s_{t+1})-\Phi(s_t),
\end{equation}
where $\Phi(s):\mathcal{S}\rightarrow\mathbb{R}$ is an arbitrary potential function~\cite{ng1999shaping}. Identifying a reward function that is not influenced by shaping is challenging because all shaped rewards equally support the expert policy. In deterministic environments, we can further define a class of state-action rewards $\tilde{r}(s,a)$ through the transition function $\mathcal{T}(s,a):\mathcal{S}\times\mathcal{A}\rightarrow \mathcal{S}$:
\begin{equation}
    \label{eq_reward_shaping_sa}
    \tilde{r}(s_t,a_t) = r(s_t,a_t)+\gamma\Phi\left(\mathcal{T}(s_t,a_t)\right) - \Phi(s_t).
\end{equation}
The policy's optimality remains unchanged if the transitions $(s_t, a_t, s_{t+1})$ in \eqref{eq_reward_shaping} are consistent with the environment dynamics, specifically if $s_{t+1} = \mathcal{T}(s_t, a_t)$. However, applying the shaped reward defined in \eqref{eq_reward_shaping_sa} to an environment with different dynamics may result in a state $s_{t+1}$ from action $a_t$ in state $s_t$ that does not align with $\mathcal{T}(s_t, a_t)$. This misalignment can influence the policy optimality, making the shaped reward dynamics-dependent. Due to the complexity of deriving a reward free from shaping, the state-action rewards generated by AIL and traditional IRL methods typically exhibit dynamics dependency.

%%%%%%%%%%%%%%%%%%%%%%%%%%%%%%%% Policy Dependency %%%%%%%%%%%%%%%%%%%%%%%%%%%%%%%%
\subsection{Policy Dependency}
\label{sec_policy_dependency}
The reward learned through AIL is also inherently dependent on policies, arising from its iterative learning framework. In AIL, the discriminator is trained to classify between expert and generated policies, while the policy itself is concurrently updated. This introduces a non-stationary training characteristic that differs from traditional supervised learning. Let the sequence of policies trained during the GAIL process be denoted as $\{\pi_1,\pi_2,\ldots,\pi_T\}$, where $\pi_t$ represents the policy at iteration $t$. The discriminator faces a continual learning challenge: at each iteration, it must discriminate the current policy $\pi_t$ from the expert policy. This scenario is similar to the neural network problem of \textit{catastrophic forgetting}~\cite{mccloskey1989catastrophic,kirkpatrick2016overcoming}, where a network loses information about earlier learned tasks as it learns new ones. In AIL, the discriminator's ability to differentiate earlier policies $\pi_1, \pi_2, \ldots, \pi_{t-1}$ diminishes as it better adapts to the current policy $\pi_t$. The discriminator will better distinguish the policy learned in the same iteration, referred to as the \textit{compatible policy}. For example, the discriminator trained at the $t$-th iteration can better distinguish the compatible policy $\pi_t$. Conversely, it struggles with \textit{incompatible policies}, those significantly divergent from the current policy, which may be entirely new or effectively forgotten.

Due to this problem, the discriminator's capacity to provide accurate rewards is constrained to a narrow spectrum of policies similar to the compatible policy. We term this restriction as policy dependency. When GAIL training converges, the discriminator becomes finely tuned to a policy closely resembling the expert policy, but this fine-tuning comes at the expense of losing the ability to differentiate the low-performance policies at early stages. If we attempt to train a new policy from scratch using this discriminator, the randomly initialized policy might receive incorrect rewards due to the discriminator's diminished capacity to differentiate it from the expert policy. The wrong reward signals could misdirect the policy learning, steering it away from achieving optimal expert behavior.

%%%%%%%%%%%%%%%%%%%%%%%%%%%%%%%% Policy Dependency %%%%%%%%%%%%%%%%%%%%%%%%%%%%%%%%

\section{Dynamic-Agnostic Discriminator Ensemble}
In this section, we introduce Dynamics-Agnostic Discriminator-Ensemble Reward Learning (\ours), which addresses the limitations previously identified in the reward learned by AIL. \ours builds upon the foundational framework of GAIL, operating with a policy $\pi(a | s; \varphi)$, where $\varphi$ represents the policy parameters, and a discriminator $D(\cdot; \psi)$, parameterized by $\psi$. Both components are trained iteratively through adversarial processes as outlined in Section~\ref{sec_preliminaries}. To address the challenges discussed in Section~\ref{sec_matters}, \ours implements specific modifications to the learning paradigm of the discriminator and the formulation of the reward function. In the subsequent parts, we will elaborate on the rationale behind these adjustments and detail their implementation.

%%%%%%%%%%%%%%%%%%%%%%%% Dynamics-Agnostic Discriminator Learning %%%%%%%%%%%%%%%%%%%%%%%%%
\subsection{Dynamics-Agnostic Discriminator Learning}
\label{sec_discriminator_learning}
\begin{figure*}[ht]
\centering 
\includegraphics[width=1.0 \linewidth]{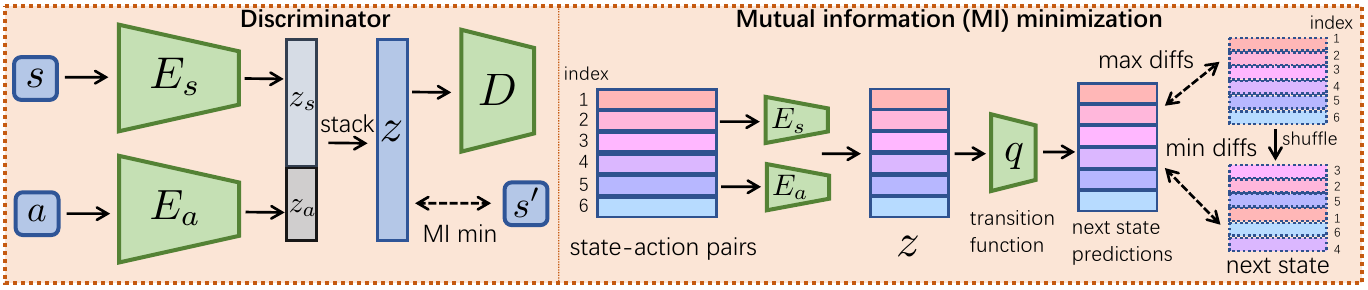}
\caption{The framework of dynamics-agnostic discriminator learning by mutual information minimization. The input of the discriminator comprises embeddings of the state-action pairs, generated by a state encoder and an action encoder. The optimization target is ensure that these embeddings have minimal minimum mutual information with the next state. The optimization objective is maximizing the prediction errors of a transition model while minimizing the prediction errors associated with incorrect labels.} 
\label{fig_framework}  
\end{figure*}
\eqref{eq_reward_shaping_sa} demonstrates that introducing a transition function $\mathcal{T}(s,a)$ allows a state-action reward $r(s,a)$ to be shaped by any potential function $\Phi(s)$. While this reward shaping preserves policy optimality in environments with trainsition dynamics $\mathcal{T}$, it can alter policy optimality in environments with different transition dynamics. Therefore, the reward shaping framework outlined in \eqref{eq_reward_shaping_sa} makes the reward coupled with the training dynamics, limiting its applicability across varied environment dynamics.

In AIL, the discriminator is trained without explicit dynamics information. Consequently, the construction of \eqref{eq_reward_shaping_sa} must occur internally and implicitly; for instance, the next state $s'$ might be forecasted in a latent space through an internally derived transition model. To prevent the discriminator from learning a shaped reward, either implicitly or explicitly, we propose to modify its input domain from the state-action space to an embedding space $\mathcal{Z}$. By minimizing the mutual information (MI) between the embeddings and the next state, we ensure that predicting the next states through the embeddings is impossible. This prevents the discriminator from completing the reward shaping construction process, ensuring that the reward model remains dynamics-agnostic and transferable across different dynamics.

\Cref{fig_framework} illustrates our proposed framework for dynamics-agnostic discriminator learning, which comprises two encoders: a state encoder $E_s(s): \mathcal{S} \rightarrow \mathcal{Z}_s$ and an action encoder $E_a(a): \mathcal{A} \rightarrow \mathcal{Z}_a$. These encoders map the state and action to their respective embeddings, $z_s$ and $z_a$. The discriminator $D(z): \mathcal{Z}_s \times \mathcal{Z}_a \rightarrow \mathbb{R}$ then processes the concatenated embeddings $z = [z_s, z_a]$. Unlike \cite{peng2019vdb}, we avoid using a universal encoder that takes the concatenation of state and action as input. This choice is motivated by the concern that a universal encoder may inadvertently incorporate dynamics-related information into the embeddings.
 
The mutual information between $s'$ and $z$ can be written as
\[
I(z;s') = \mathbb{E}_{p(z,s')}[\log p(s'| z)] - \mathbb{E}_{p(s')}[\log(p(s'))]. 
\]
Minimizing $I(z;s')$ directly is intractable because  $p(s'| z)$ is unknown. Instead, we can minimize an upper bound of $I(z;s')$. We use a recently proposed upper bound of $I(z;s')$, the variational contrastive log-ratio upper bound (vCLUB)~\cite{cheng2020vclub}. 
\begin{theorem}[Variational Contrastive Log-Ratio Upper Bound~\cite{cheng2020vclub}]
\label{thm_vclub}
Let $q(s'| z;\theta)$ be a variational approximation of $p(s'| z)$ with parameter $\theta$.  Denote $q(z,s';\theta)=q(s'| z;\theta)p(z)$. If 
\begin{equation}
\label{eq_club_constraint}    
D_{KL}(p(z,s') \Vert q(z,s';\theta))\leq D_{KL}(p(s')p(z)\Vert q(z,s';\theta)),
\end{equation}
then $I(z;s') \leq I_\text{vCLUB}(z;s')$, where
\begin{equation}
\label{eq_club_upper_bound}    
\resizebox{0.45\textwidth}{!}{$
I_\text{vCLUB}= \mathbb{E}_{p(z,s')}\left[\log q(s'| z;\theta)\right]-\mathbb{E}_{p(z)}\mathbb{E}_{p(s')}\left[\log q(s'| z;\theta)\right].
$}
\end{equation}
\end{theorem}
For a detailed proof, please refer to \ref{supp_theorem_proof} or \cite{cheng2020vclub}.
Theorem~\ref{thm_vclub} provides a method to minimize the MI by minimizing a tractable MI upper bound. We introduce a transition network $q(s'| z;\theta)$ parameterized by $\theta$ to approximate $p(s'| z)$. Given a sample set $\{(s'^i,z^i)\}_{i=1}^N\sim p(z,s')$, the vCLUB from \eqref{eq_club_upper_bound} can be empirically estimated as
\begin{equation}
    \label{eq_empirical_club}
    \hat{I}_\text{vCLUB} = \sum_{i=1}^N\frac{\log q(s'^i| z^i;\theta)}{N} - \sum_{i=1}^N\sum_{j=1}^N\frac{\log q(s'^j| z^i;\theta)}{N^2}.
\end{equation}
Minimizing \eqref{eq_empirical_club} involves two main objectives: decreasing the first term to optimize the embedding $z$ to prevent $q$ from accurately inferring $s'$, and increasing the second term to ensure that $q$ infers incorrect $s'$ from $z$. The process of mutual information minimization is illustrated in the right section of \cref{fig_framework}.

To reduce computational complexity, we avoid the exhaustive computation of the second term in \eqref{eq_empirical_club}. Instead, we approximate this term using the prediction error of $q$ on a permuted dataset, where the indices of $s'$ are randomly shuffled. This effectively simulates the scenario where $q$ infers incorrect $s'$ from $z$.

The combination of $E_s$ and $E_a$ is defined as $E(s,a;\phi): \mathcal{S} \times \mathcal{A} \rightarrow \mathcal{Z}_s \times \mathcal{Z}_a$, parameterized by $\phi$. The loss function for $E(s,a;\phi)$, given a dataset $\mathcal{B} = {s^i, a^i, s'^i}$, is formulated as follows:
\begin{equation}
    \label{eq_loss_encoder}
    \begin{aligned}
    \mathcal{L}_E(\mathcal{B},\phi)&=\mathbb{E}_{s,a,s'\sim \mathcal{B}}[\log q(s'| E(s,a;\phi);\theta)] \\
    &~~~~~~~~- \mathbb{E}_{s,a,\tilde{s}'\sim \mathcal{B}_\text{suf}}[\log q(\tilde{s}'| E(s,a;\phi);\theta)],
    \end{aligned}
\end{equation}
where $\mathcal{B}_\text{suf}$ represents the dataset obtained by shuffling the indices of $s'$ within $\mathcal{B}$, and $(s,a,s')\sim\mathcal{B}$ indicates sampling uniformly from $\mathcal{B}$.

Furthermore, \eqref{eq_empirical_club} serves as a valid upper bound for MI only when \eqref{eq_club_constraint} is satisfied. The left-hand side of \eqref{eq_club_constraint} can be simplified as follows:
\begin{equation}
\label{eq_kl_simplication}
\begin{aligned}
&~~~~~D_{KL}(p(z,s') \Vert q(z,s';\theta)) \\
&= \mathbb{E}_{p(z,s')}\left[\log p(z,s') \right] - \mathbb{E}_{p(z,s')}\left[\log q(z,s';\theta) \right]\\
&= \mathbb{E}_{p(z,s')}\left[\log p(s'| z) \right] - \mathbb{E}_{p(z,s')}\left[\log q(s'| z;\theta) \right].\\
\end{aligned}
\end{equation}
In \eqref{eq_kl_simplication}, the first term is constant with respect to $q$, indicating that $q$ should maximize the second term to satisfy \eqref{eq_club_constraint}. Consequently, the loss for $q(s'| z;\theta)$ is defined as:
\begin{equation}
    \label{eq_loss_transition}
    \mathcal{L}_q(\mathcal{B}, \theta)=-\mathbb{E}_{s,a,s'\sim \mathcal{B}}[\log q(s'| E(s,a;\phi);\theta)].
\end{equation}
Comparing \eqref{eq_loss_encoder} and \eqref{eq_loss_transition}, we can find that the encoder and the transition network are trained adversarially. However, such training may result in an uninformative embedding $z$, such as a completely random variable. To address this, we train the encoder in conjunction with a discriminator, ensuring that $z$ contains information about the reward while minimizing dynamics-related information. Consequently, we aim for the embedding $z$ to exclusively encapsulate reward information, with minimal dynamics-related information. Ultimately, the discriminator loss for \ours is given by:
\begin{equation}
    \label{eq_modified_d_loss}
    \resizebox{0.45\textwidth}{!}{$
    \begin{aligned}
    \mathcal{L}(\mathcal{B}^\pi,\mathcal{B}^E, \psi, \phi) &=\mathbb{E}_{s,a\sim\mathcal{B}^E}[\log (D(E(s,a;\phi);\psi))] \\
    &+ \mathbb{E}_{s,a\sim\mathcal{B}^\pi}[\log(1-D(E(s,a;\phi);\psi))] \\&+ \eta\mathcal{L}_E(\mathcal{B}^\pi \cup\mathcal{B}^E,\phi),
    \end{aligned}$
    }
\end{equation}
where $\eta$ is a regularization factor.
%%%%%%%%%%%%%%%%%%%%%%%% Dynamics-Agnostic Discriminator Learning %%%%%%%%%%%%%%%%%%%%%%%%%

%%%%%%%%%%%%%%%%%%%%%%%%%%%%%%% Discriminator Ensemble %%%%%%%%%%%%%%%%%%%%%%%%%%%%%%%%%
\subsection{Discriminator Ensemble}
\label{sec_disc_ensemble}
In this part, we address the policy dependency problem by analyzing the properties of the discriminator learned through AIL. We introduce a special implementation of GAIL, termed Online Gradient Descent-GAIL (OGD-GAIL), as presented in Alg.~\ref{alg_OGD}. In line~\ref{alg_ogd_update_d}, $\Pi_{(0,1)^{|\mathcal{S}|\times|\mathcal{A}|}}(\cdot)$ denotes constraining each entry of the value in the brackets to the range of $(0,1)$~\cite{bubeck2015convex_optimization}, which makes $D_{t+1}$ legal. Alg.~\ref{alg_OGD} assumes the finite $\mathcal{S}$ and $\mathcal{A}$. Each entry of $D_t\in(0,1)^{|\mathcal{S}|\times|\mathcal{A}|}$ corresponds to the discriminator output value for each state-action pair in $\mathcal{S}\times\mathcal{A}$. Let the mean output of the discriminators in all iterations be
\begin{equation}
\label{eq_mean_output_of_d}
    \overline{D}(s,a) = \frac{1}{T}\sum_{t=1}^T D_{t}(s,a),~~~~\forall s\in\mathcal{S},a\in\mathcal{A}.
\end{equation}
We can find the average loss, $\frac{1}{T}\sum_{t=1}^T\mathcal{L}(\overline{D}, \pi_t)$, for all past policies can be bounded under some mild assumptions, which we summarize in Theorem~\ref{thm_emsemble}. 
\begin{algorithm}[t]
\caption{Online Gradient Descent-GAIL} \label{alg_OGD}
\KwIn{Step sizes $\{\omega_1,\omega_2,\dots,\omega_T\}$ and expert demonstrations $\mathcal{B}^E$.}
Initialize discriminator $D_{1}$\;
\For {t $=1,2,\cdots,T$}
{
Update the policy: 
$\pi_t \leftarrow \arg\max_{\pi} \mathcal{L}(D_t, \pi) $\;\label{alg_ogd_update_g}
Update the discriminator: $D_{t+1} \leftarrow \Pi_{(0,1)^{|\mathcal{S}|\times|\mathcal{A}|}} \left( D_t - \omega_t\nabla_{D_t} \mathcal{L}(D_t, \pi_t)\right)$\;
\label{alg_ogd_update_d}
}
\end{algorithm}

\begin{theorem}[Discriminator Ensemble Upper Bound]
\label{theorem_main_conclusion}
Consider finite $\mathcal{S}$ and $\mathcal{A}$, if $\max_{\pi\in\{\pi_t| t\in[1,T]\}}\| \nabla_D \mathcal{L}(D, \pi_t) \|_2 \leq G$, Alg.~\ref{alg_OGD} with step sizes $\{ \omega_t = \frac{\Omega}{G\sqrt{t}}, t=1,2,\cdots,T \}$ will achieve the following guarantee for all $T \geq 1$:
\begin{equation}
    \frac{1}{T}\sum_{t=1}^T\mathcal{L}(\overline{D},\pi_t)  \leq \frac{1}{T}\min_D\sum_{t=1}^T \mathcal{L}(D,\pi_{t}) + \frac{3\overline{\Omega} G}{2\sqrt{T}},
\end{equation}
where $\Omega$ is a hyperparameter and $\overline{\Omega}=\max\{\frac{\|D_T-D^\star\|_2^2}{\Omega}, \Omega\}$.
\label{thm_emsemble}
\end{theorem}
\begin{proof}
Please refer to \ref{proof_thm_emsemble} for detailed proof. The proof sketch is to show that $\mathcal{L}(D,\cdot)$ is convex first, and then the regret analysis follows the standard regret analysis for online gradient descent.
\end{proof}
Due to the upper bound of $\|D_T-D^\star\|_2^2$ being a constant, specifically $|\mathcal{S}|\times|\mathcal{A}|$, the term $\frac{3\overline{\Omega} G}{2\sqrt{T}}$ converges to $0$ as $T\rightarrow \infty$. Consequently, Theorem~\ref{thm_emsemble} suggests that as $T$ increases, the mean discriminator $\overline{D}$ converges to an optimal discriminator with minimal loss across all historical policies. OGD-GAIL (Alg.~\ref{alg_OGD}) operates as a minimax optimization, with $\pi_t$ consistently responding to the current $D_t$. This process follows a ``no-regret vs. best response'' paradigm, indicating that the average model, represented by $\overline{D}$, forms an $\epsilon$-Nash equilibrium, as stated in Theorem~3 of \cite{cfr_vs_br:12}, where $\epsilon$ is the upper bound on average regret. On the other hand, based on Theorem~4 in \cite{cfr_vs_br:12}, if a single $D$ is randomly chosen from the $T$ discriminators, it constitutes a $(T\epsilon)$-Nash equilibrium with a probability of $1/T$. However, employing an ensemble of all discriminators significantly improves the convergence towards a Nash equilibrium.

Thus, retaining all discriminators obtained during the training process and constructing a discriminator ensemble enhances discriminative capabilities, enabling better distinction among all historical policies and mitigating the policy dependency issue. Notably, previous studies have also introduced generator or discriminator ensembles in the context of generative adversarial networks~\cite{goodfellow2020gan}, yielding convergence guarantees~\cite{arora2017quilibrium_gan,hsieh2019mixednash,Aung2022dogan}.

%%%%%%%%%%%%%% How DARL do %%%%%%%%%%%%%%%%%

Building upon the insights mentioned above, \ours attempts to address the policy dependency issue by leveraging an ensemble of historical discriminators. To achieve this, \ours initializes an empty buffer, $\mathcal{C}$, and populates it with learned discriminators at fixed intervals of every $H$ iterations. Upon completion of the imitation algorithm, \ours constructs the reward function using $\mathcal{C}$. The reward for any state-action pair is determined from the ensemble's mean discriminator output, computed as:
\begin{equation}
\label{eq_reward_ensemble_pre}
r^{\text{origin}}_E(s,a) = -\log(1 - \sum_{i=1}^{|\mathcal{C}|} D_i(s,a) / |\mathcal{C}| ),
\end{equation}
where $D_i(s,a)$ denotes the $i$-th discriminator in $\mathcal{C}$. Our theoretical results, presented in Theorem~\ref{theorem_main_conclusion}, indicate that the integration of all past discriminators within the ensemble enables effective distinction of all observed policies, thus theoretically resolving the policy dependency issue. Furthermore, with multiple discriminators, the ensemble includes those compatible with policies from various training stages. Consequently, at different iterations during policy training, there exists a discriminator within the ensemble that offers relatively accurate rewards, guiding the policy effectively.

\noindent{\textbf{Implementation Modifications.}} Based on $r^{\text{origin}}_E(s,a)$, we further introduce two practical modifications to balance the outputs of the discriminators in $\mathcal{C}$ and prevent them from being over-exploited. 

The first adjustment addresses the variability in the output ranges of discriminators learned at different training stages. Early-stage discriminators, which often struggle to differentiate between expert and learner behaviors, typically output values around $0.5$. In contrast, discriminators that have reached convergence output values close to $1.0$ for expert data and near $0.0$ for non-expert data. This disparity can cause later-stage discriminators to dominate the variation range of the ensemble reward function, leading to a policy that prioritizes optimizing the outputs of later discriminators while neglecting earlier ones. To mitigate this, we normalize the output of each discriminator to ensure a uniform output range across all discriminators in $\mathcal{C}$. This normalization ensures a balanced contribution from each discriminator to the final ensemble reward.
\begin{equation}
    \label{eq_reward_transform_trick_normalize}
    \begin{aligned}
D^\text{norm}_i(s,a) = \tilde{D}_i(s,a)\left(D^\text{max} - D^\text{min}\right) + D^\text{min},
\end{aligned}
\end{equation}
where $\tilde{D}_i(s,a)$ represents the normalized value of the $i$-th discriminator for a given state-action pair $(s, a)$, and is computed as:
\begin{equation}
\label{eq_reward_transform_trick_normalize_supp}
\tilde{D}_i(s,a) = \texttt{clip}_{0,1}\left(\frac{D_i(s,a) - D^{\text{learner}}_i}{D^{\text{expert}}_i- D^{\text{learner}}_i}\right).
\end{equation}
Here, $\texttt{clip}_{[0,1]}(\cdot)$ denotes the operation of clipping its argument to the range $[0,1]$. $D^{\text{expert}}_i$ and $D^{\text{learner}}_i$ represent the average outputs of discriminator $D_i$ for expert and learner data, respectively. $D^\text{min}=\min_i D^{\text{learner}}_i$ and $D^\text{max}=\max_i D^{\text{expert}}_i$ are the minimum and maximum discriminator outputs observed during training, respectively. By normalizing $D_i$ with \eqref{eq_reward_transform_trick_normalize_supp}, we align the output range of each discriminator to $[D^{\text{min}}, D^{\text{max}}]$, as specified in \eqref{eq_reward_transform_trick_normalize}.

The second modification, formalized in \eqref{eq_reward_transform_trick_clip}, involves clipping the normalized discriminator outputs, $D_i^\text{norm}$, at a threshold $c$. This modification is introduced to prevent the RL agent from over-exploiting individual discriminators. Due to policy dependency, a single discriminator's output may be unreliable. The clipping technique ensures that the RL agent does not excessively focus on a single discriminator or a subset of discriminators, thereby forcing the agent to optimize the collective outputs of all discriminators and reducing the risk of training bias.
\begin{equation}
    \label{eq_reward_transform_trick_clip}
D^\text{norm\_clip}_i(s,a) = \min(D^\text{norm}_i(s,a), c).
\end{equation}
Finally, the reward derived from the ensemble model is defined as:
\begin{equation}
\label{eq_reward_enesmble}
r_E(s,a) = -\log\left(1 - \sum_{i=1}^{|\mathcal{C}|} D^\text{norm\_clip}_i(s,a)/ |\mathcal{C}| \right).
\end{equation}

%%%%%%%%%%%%%%%%%%%%%%%%%%%%%%% Discriminator Ensemble %%%%%%%%%%%%%%%%%%%%%%%%%%%%%%%%%

\subsection{The \ours Algorithm}
We provide an overview of the training process for \ours in Alg.~\ref{alg_main_algorithm}. \ours is built upon the framework of GAIL, incorporating two encoders and a transition network designed to eliminate dynamics-related information from the input of the discriminator. This is accomplished by minimizing an upper bound on MI. To ensure this upper bound is effective, we optimize the parameters of the transition network using both historical and expert data, as detailed in line\ref{alg_main_line_ref2} of Alg.~\ref{alg_main_algorithm}. To reduce policy dependency, \ours periodically saves the historical discriminators learned during training, as highlighted in line~\ref{alg_main_line_ref3}. To align practically with Alg.~\ref{alg_OGD} and the theoretical results in Theorem~\ref{thm_emsemble}, we adopt a relatively large value for g\_steps, for example, $5$ or $10$, to approximate the $\arg\max$ operation in line~\ref{alg_ogd_update_g} of Alg.~\ref{alg_OGD}. Additionally, we set d\_steps to $1$, mirroring the one-step discriminator update in line~\ref{alg_ogd_update_d} of Algorithm~\ref{alg_OGD}.

\begin{algorithm}[t]
\caption{Dynamics-Agnostic Discriminator-Ensemble Reward Learning (\ours)} \label{alg_main_algorithm}
\KwIn{Expert data $\mathcal{B}^E$; policy $\pi(a| s;\varphi)$; discriminator $D(z;\psi)$; encoder $E(s,a;\phi)$; transition $q(s,a;\theta)$; policy, discriminator, and transition updating steps, g\_steps, d\_steps, and transition\_steps; discriminator backup interval $H$.}
Initialize empty disc. buffer $\mathcal{C}$ and memory buffer $\mathcal{R}$\;

\For {t $\in [1,$ max\_iterations$]$}
{
\For{step\_g $=1,\dots,$ g\_steps}{
    Sample data $\mathcal{B}^\pi$ with $\pi(a| s;\varphi)$, insert $\mathcal{B}^\pi$ to $\mathcal{R}$\;
    Calculate the reward for the data in $\mathcal{B}^\pi$ with $\hat{r}(s,a)=-\log\left(1-D(E(s,a;\phi);\psi)\right)$\;
    Update $\varphi$ with data in $\mathcal{B}^\pi$ and $\hat{r}$ via PPO~\cite{schulman2017ppo}\;
}
\For{step\_d $=1,\dots,$ d\_steps}{
    Update $\psi,\phi$ by \eqref{eq_modified_d_loss} with $\mathcal{B}^\pi$ and $\mathcal{B}^E$\;\label{alg_main_line_ref1}
}
\For{step\_transition $=1,\dots,$ transition\_steps}{
    Sample data $\mathcal{B}^\pi$ from $\mathcal{R}$\;
    Update $\theta$ by \eqref{eq_loss_transition} with $\mathcal{B}^\pi$ and $\mathcal{B}^E$\;\label{alg_main_line_ref2}
}
\If{$(t-1)$ \texttt{mod} $ H==0$}
{
  Insert $(\phi,\psi)$ into $\mathcal{C}$\;\label{alg_main_line_ref3}
}
}
\KwOut{Reward represented by \eqref{eq_reward_enesmble} based on $\mathcal{C}$.}
\end{algorithm}
\section{Experiments}
\label{sec_experiment}
In this section, we conduct a series of experiments designed to investigate the following questions:
\begin{enumerate}
    \item Does the policy dependency problem exist in current AIL methods? (\cref{fig_exp_policy_denpendency_halfcheetah})
    \item How does \ours solve the policy and dynamics dependency problems? (\cref{fig_exp_fix_policy_denpendency_halfcheetah,fig_training_process_grid_view,fig_dynamics_gravity_curve_fix_ret})
    \item Can \ours learn a reward function consistent with the true reward function? (\cref{tab_reward_consistency})
    \item What performance can a policy achieve with the rewards learned by \ours in transfer scenarios? (\cref{fig_main_comparison})
    \item Can \ours handle harder problems? (\cref{fig_hard_task,fig_hard_task_humanoid,fig_rampti_task,tab_consistency_humanoid})
    \item What are the effects of the reward modification techniques and the hyper-parameters? (\cref{fig_ablation,fig_ablation_reward_transform,fig_ablation_c_ant})
    \item What have the encoders learned? (\cref{fig_response_tsne})
\end{enumerate}

\subsection{Setup}
We evaluate \ours across five MuJoCo~\cite{todorov2012mujoco} tasks, i.e., \texttt{Hopper}, \texttt{HalfCheetah}, \texttt{Walker2d}, \texttt{Ant}, and \texttt{Humanoid}. For each environment, we train a policy from scratch using PPO~\cite{schulman2017ppo} and utilize the deterministic learned policy to generate trajectories as expert demonstrations. Subsequently, we apply each IRL/AIL method to these demonstrations to derive the reward function, which we save at the last iteration for downstream tasks.

To construct the transfer tasks\footnote{The expert policy, demonstrations, and reward function are all trained/collected in the original environment without dynamics disturbances.}, we change the dynamics parameters of the environments such as \texttt{gravity} and \texttt{dof\_damping}. \texttt{Gravity} affects how objects fall and interact, simulating different weight conditions, while \texttt{dof\_damping} impacts the stability and responsiveness of joints, controlling the fluidity of motion. As designed in some meta-RL literature~\cite{rakelly2019pearl,peng2018sim2real,luo2022escp}, each transfer task consists of 20 environments with different dynamics parameters, sampled independently and uniformly from a given distribution. We expect that the learned reward functions can be robust enough to provide accurate reward signals across various environment dynamics. Additionally, for \texttt{Ant}, we also construct a transfer task \texttt{DisableAnt} by disabling half of its legs, which is the same as in previous works~\cite{ni2020firl,fu2017airl,zeng2022mlirl}.

We use PPO~\cite{schulman2017ppo} as the RL algorithm in \ours and the variants of GAIL. Each experiment and each method is run with $6$ distinctive seeds. 
All experiments were conducted on a Mac Studio equipped with an Apple M1 Ultra CPU and 128GB of RAM. The number of the training iterations is $100$ ($200$ for \texttt{Humanoid}). The discriminator backup interval $H$ is set to $1$ ($10$ for \texttt{Humanoid}), resulting in a discriminator ensemble size of $100$ or $20$, respectively. More details of the experiments and algorithm implementations can be found in \ref{supp_experiment_details}. 

\noindent{\textbf{Baseline methods.}} We compare \ours with $3$ kinds of baselines. The first category of methods includes variations of GAIL, each of which attempts to alleviate either policy or dynamics dependency. 
\begin{itemize}
    \item \textit{GAIL}~\cite{ho2016gail}: The original GAIL method.
    \item \textit{GAIL-B} (GAIL-Buffer): Enhances GAIL by integrating a replay buffer to retain all data sampled by the policy, thus reducing policy dependency through comprehensive historical data preservation.
    \item \textit{GAIL-DAC}: Builds upon GAIL-B by incorporating an absorbing state mechanism to eliminate terminal state reward bias, as proposed in~\cite{kostrikov2019dac}.
    \item \textit{GAIL-E} (GAIL-Ensemble): Utilizes a collective of historical discriminators, leveraging the averaged output to formulate the reward (\eqref{eq_reward_enesmble}).
    \item \textit{VAIL}~\cite{peng2019vdb}: Introduces a variational approach to minimize mutual information between the input and an internal layer's output of the discriminator.
    \item \textit{AIRL-SA}~\cite{fu2017airl}: Adjusts the discriminator architecture to facilitate the learning of an unshaped reward and represents the reward function as $\hat{r}_\text{AIRL}(s,a)=\log(D(s,a))-\log(1-D(s,a))$.
    \item \textit{AIRL-SO}~\cite{fu2017airl}: The state-only variant of AIRL-SA, specializing in learning rewards dependent solely on the state, enhancing transferability across different dynamics.
\end{itemize}
The second category of baselines contains the deep IRL methods. 
\begin{itemize}
    \item \textit{$f$-IRL}~\cite{ni2020firl}: Employs state marginal matching under $f$-divergence to infer a stationary reward function, recovering a state-only reward to facilitate reward recovery. 
    \item \textit{MCE-IRL}~\cite{ziebart2008maxent,zibart2010mceirl}: Represents a classic approach in maximum entropy IRL, using a neural network to estimate the reward function. For our experiments, we adapt the implementation from Ni et al.~\cite{ni2020firl}.
    \item \textit{IQ-Learn} (Inverse soft-Q Learning)~\cite{garg2021iql}: Adopts a non-adversarial stance in the imitation learning framework, also applicable to IRL problems.
\end{itemize}

For both $f$-IRL and IQ-Learn, policies are optimized using SAC~\cite{haarnoja2018sac} following their respective official implementations. Additionally, MCE-IRL is built upon the $f$-IRL codebase and also uses SAC for policy training. The third category of baselines includes other non-imitation learning methods.
\begin{itemize}
 \item \textit{transferred policy}, i.e., the transfer performance of the converged policy learned  by GAIL in the original environment. 
 \item \textit{oracle}, training policies with the true reward via PPO~\cite{schulman2017ppo}. 
\end{itemize}

\subsection{Policy Dependency Issue in GAIL}
\label{sec_policy_dependency_problem}

\begin{figure}[htb]
     \centering
     \begin{subfigure}[b]{0.45 \linewidth}
         \centering
         \includegraphics[width=1.0 \linewidth]{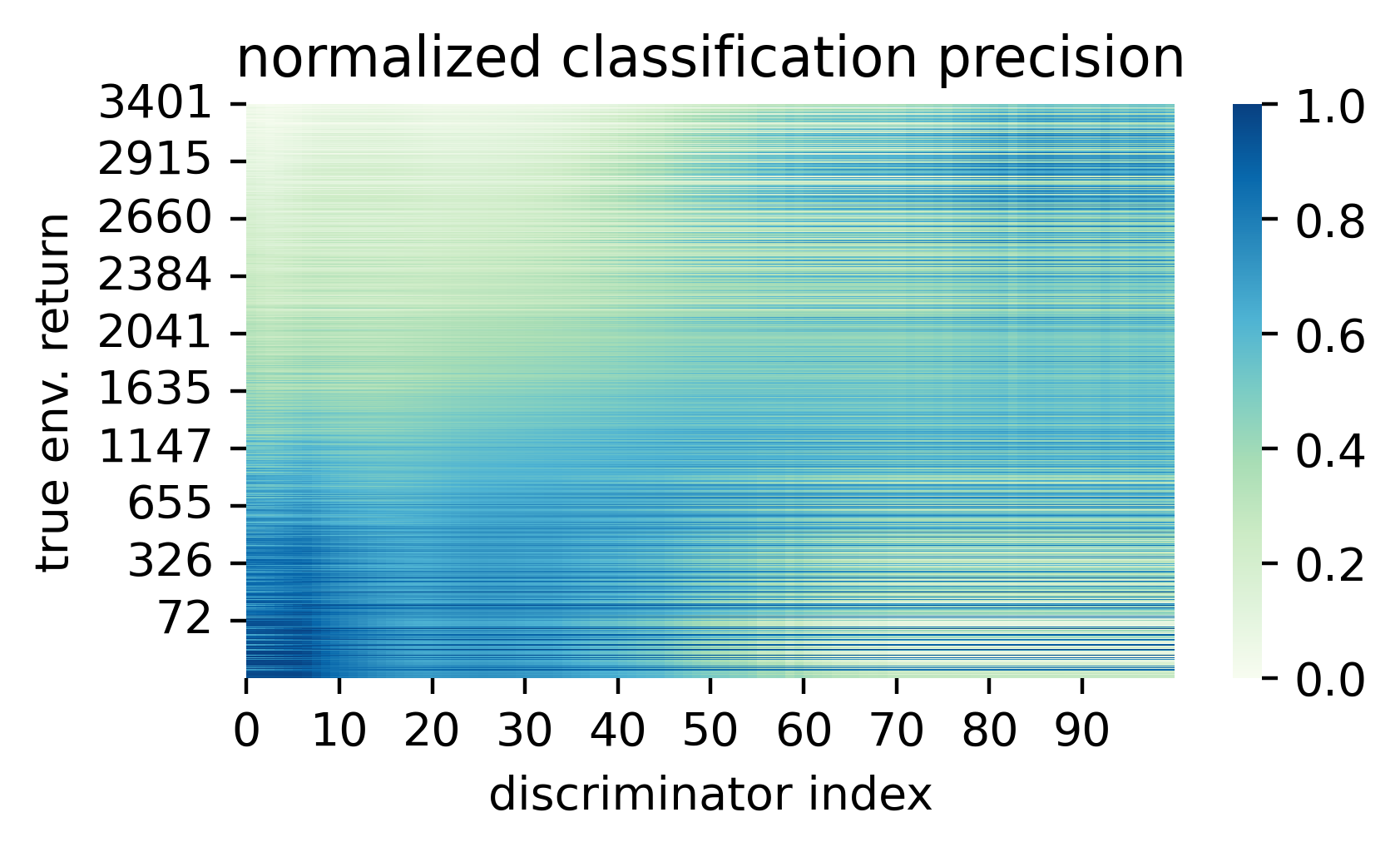}
\caption{Classification precision of the discriminators.} 
         \label{fig_policy_dependency_classification_precision}
     \end{subfigure}
     % \hfill
     \begin{subfigure}[b]{0.45\linewidth}
         \centering
         \includegraphics[width=1.0 \linewidth]{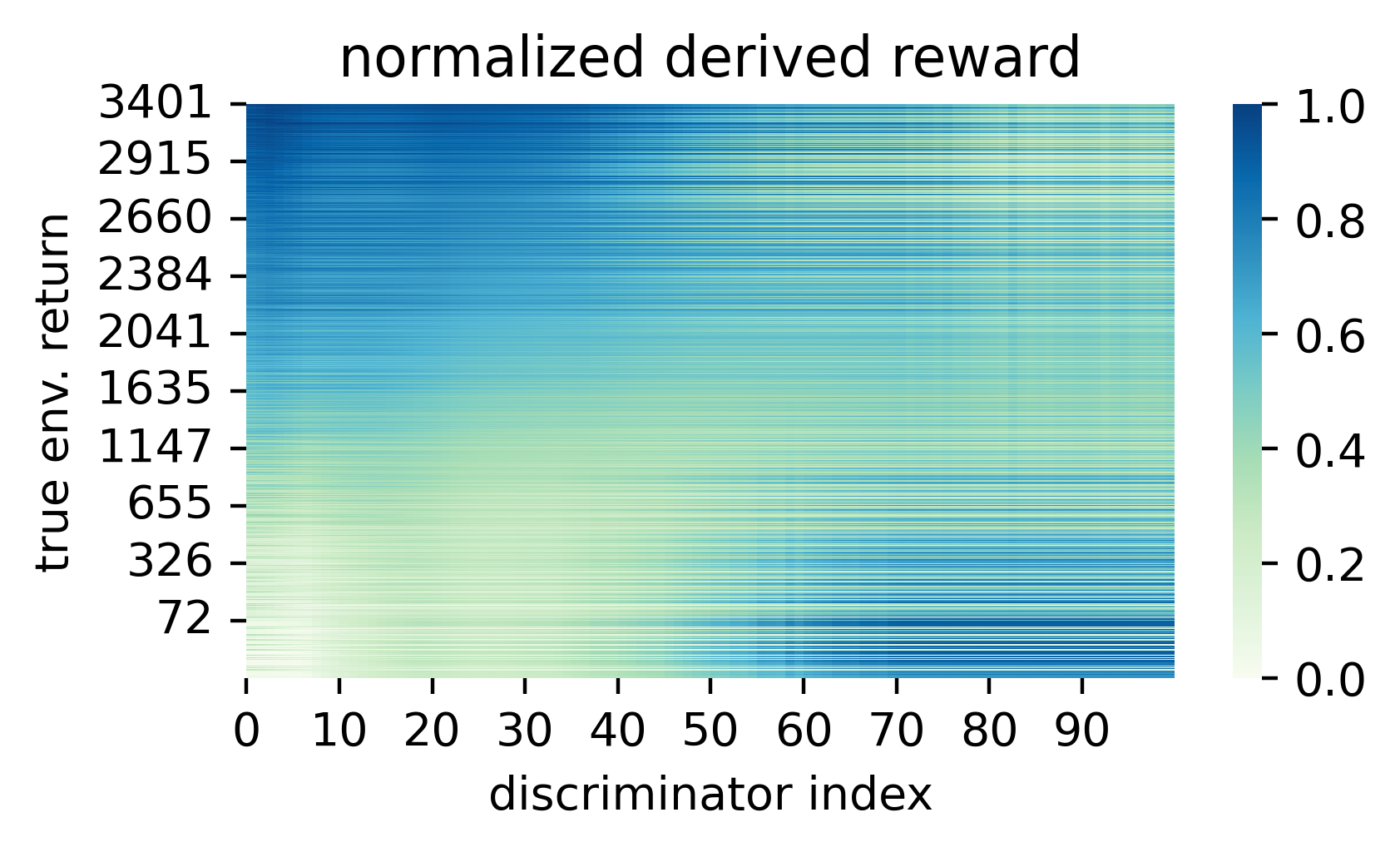}
\caption{Normalized rewards derived by the discriminators.} 
         \label{fig_policy_dependency_derived_reward}
     \end{subfigure}
        \caption{Classification precision and derived reward of the discriminators learned successively in a GAIL training process. The precision and the derived reward are evaluated on a set of data with various returns.  }
        \label{fig_exp_policy_denpendency_halfcheetah}
\end{figure}

\begin{table*}[tbp]
    \centering
    \caption{Reward consistency $\pm$ standard error in transferred environments. The changed dynamics parameter is \texttt{gravity}. 
    }
    \label{tab_reward_consistency}
    \resizebox{1.0\textwidth}{!}{
    \begin{tabular}{l|r@{~$\pm$~}lr@{~$\pm$~}lr@{~$\pm$~}lr@{~$\pm$~}lr@{~$\pm$~}lr@{~$\pm$~}lr@{~$\pm$~}lr@{~$\pm$~}l}\toprule
& \multicolumn{2}{c}{DARL} & \multicolumn{2}{c}{GAIL} & \multicolumn{2}{c}{GAIL-B} & \multicolumn{2}{c}{GAIL-DAC} & \multicolumn{2}{c}{GAIL-E} & \multicolumn{2}{c}{VAIL} & \multicolumn{2}{c}{AIRL-SA} & \multicolumn{2}{c}{AIRL-SO}\\\midrule
Ant-v2 & $ \mathbf{0.97} $ & $ \mathbf{0.00} $ & $0.51$& $0.09$ & $0.39$& $0.11$ & $0.55$& $0.10$ & $0.88$& $0.02$ & $0.43$& $0.04$ & $0.38$& $0.04$ & $0.50$& $0.04$\\
HalfCheetah-v2 & $ \mathbf{0.96} $ & $ \mathbf{0.01} $ & $0.50$& $0.11$ & $0.74$& $0.11$ & $0.90$& $0.01$ & $0.76$& $0.09$ & $0.88$& $0.01$ & $0.38$& $0.02$ & $0.35$& $0.05$\\
Hopper-v2 & $ \mathbf{0.94} $ & $ \mathbf{0.01} $ & $0.83$& $0.06$ & $0.78$& $0.04$ & $0.88$& $0.02$ & $0.87$& $0.02$ & $\mathbf{0.94}$& $\mathbf{0.01}$ & $0.66$& $0.06$ & $0.57$& $0.03$\\
Walker2d-v2 & $ \mathbf{0.94} $ & $ \mathbf{0.01} $ & $0.91$& $0.02$ & $0.93$& $0.02$ & $0.93$& $0.01$ & $0.87$& $0.01$ & $0.86$& $0.02$ & $0.64$& $0.06$ & $0.72$& $0.06$\\
Humanoid-v2 & $ \mathbf{0.92} $ & $ \mathbf{0.01} $ & $0.88$& $0.01$ & $0.89$& $0.01$ & $0.88$& $0.02$ & $0.58$& $0.03$ & $0.44$& $0.04$ & $0.88$& $0.02$ & $0.87$& $0.01$\\
\bottomrule
    \end{tabular}
    }
\end{table*}

In this part, we would like to verify whether the policy dependency exists in GAIL (\cref{sec_policy_dependency}). We choose \texttt{HalfCheetah} as the testbed for this experiment as there is no terminal state. The trajectory length in \texttt{HalfCheetah} is constant, making the returns dependent solely on the average rewards. Therefore, in this environment, we can directly compare the average rewards from the learned reward model with the policy performance, represented by the true environment returns. We train a policy using GAIL and save the learned discriminators at each iteration. Additionally, we train a policy from scratch in the same environment using the true reward, preserving all sample data collected during training. We denote the data at the $i$-th iteration by $\mathcal{B}_i$. For every iteration $i$ and any discriminator $D$, we calculate the classification precision 
$\mathcal{P}(\mathcal{B}_i)=1-\sum_{s,a\in\mathcal{B}_i}{D(E(s,a;\phi);\psi)}/{|\mathcal{B}_i|}.
$
The derived reward is
$\mathcal{R}(\mathcal{B}_i) = \sum_{s,a\in\mathcal{B}_i}{-\log(1-D(E(s,a;\phi);\psi))}/{|\mathcal{B}_i|}.
$
We normalize $\mathcal{P}$ and $\mathcal{R}$ using min-max normalization, ensuring their values range between $0$ and $1$. The results are presented in \cref{fig_exp_policy_denpendency_halfcheetah}.
We observe that the discriminator learned in the early stages can better classify the low-return data, as shown on the left-bottom side of \cref{fig_policy_dependency_classification_precision}. In contrast, the discriminator at convergence, depicted on the right side, struggles to distinguish the low-return data. Conversely, these discriminators achieve higher classification precision on high-return data. Consequently, the discriminator at convergence tends to reward the low-return data more than the high-return data (\cref{fig_policy_dependency_derived_reward}). Ideally, the low-return data should be easier to distinguish than the high-return data. However, the discriminator gradually forgets how to differentiate the early policies. The results presented in \cref{fig_exp_policy_denpendency_halfcheetah} illustrate the policy dependency phenomenon and confirm its existence within GAIL.

\subsection{Alleviating Policy Dependency by \ours} 

\begin{figure}[htb]
     \centering
     \begin{subfigure}[b]{0.45 \linewidth}
         \centering
         \includegraphics[width=1.0 \linewidth]{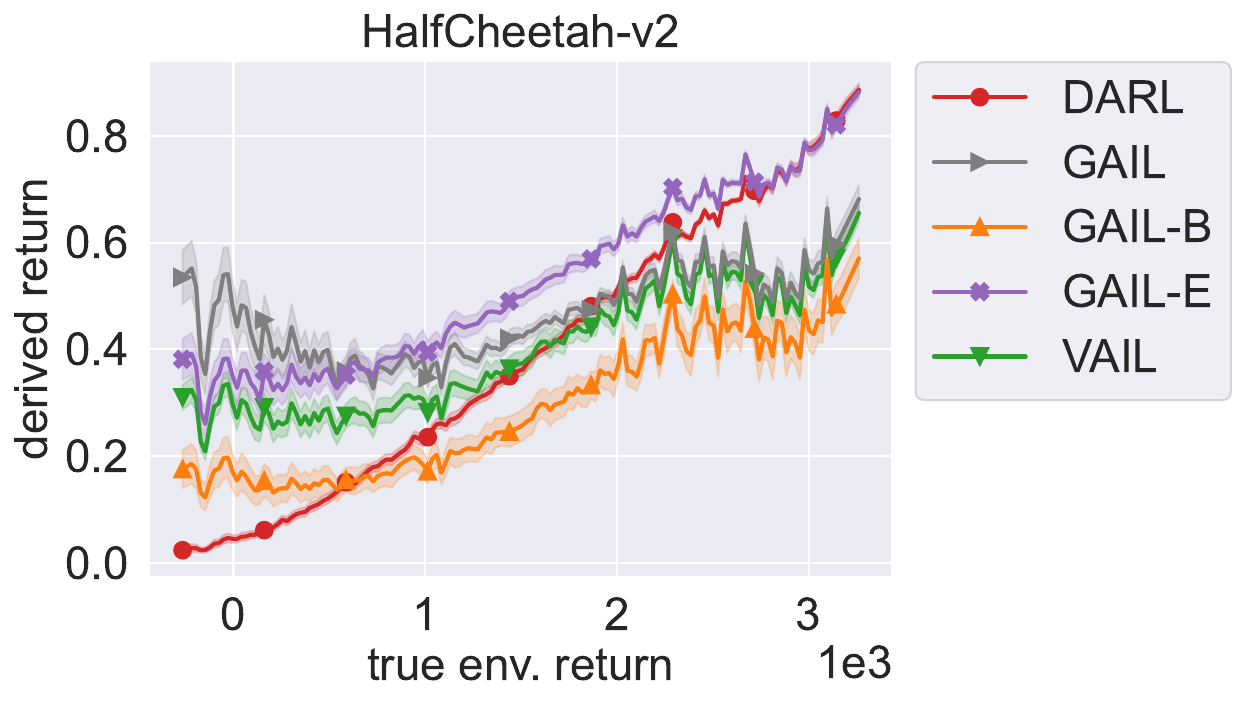}
\caption{Derived return v.s. true environment return.} 
         \label{fig_exp_fix_policy_dependency_classification_precision}
     \end{subfigure}
     % \hfill
     \begin{subfigure}[b]{0.45\linewidth}
         \centering
         \includegraphics[width=1.0 \linewidth]{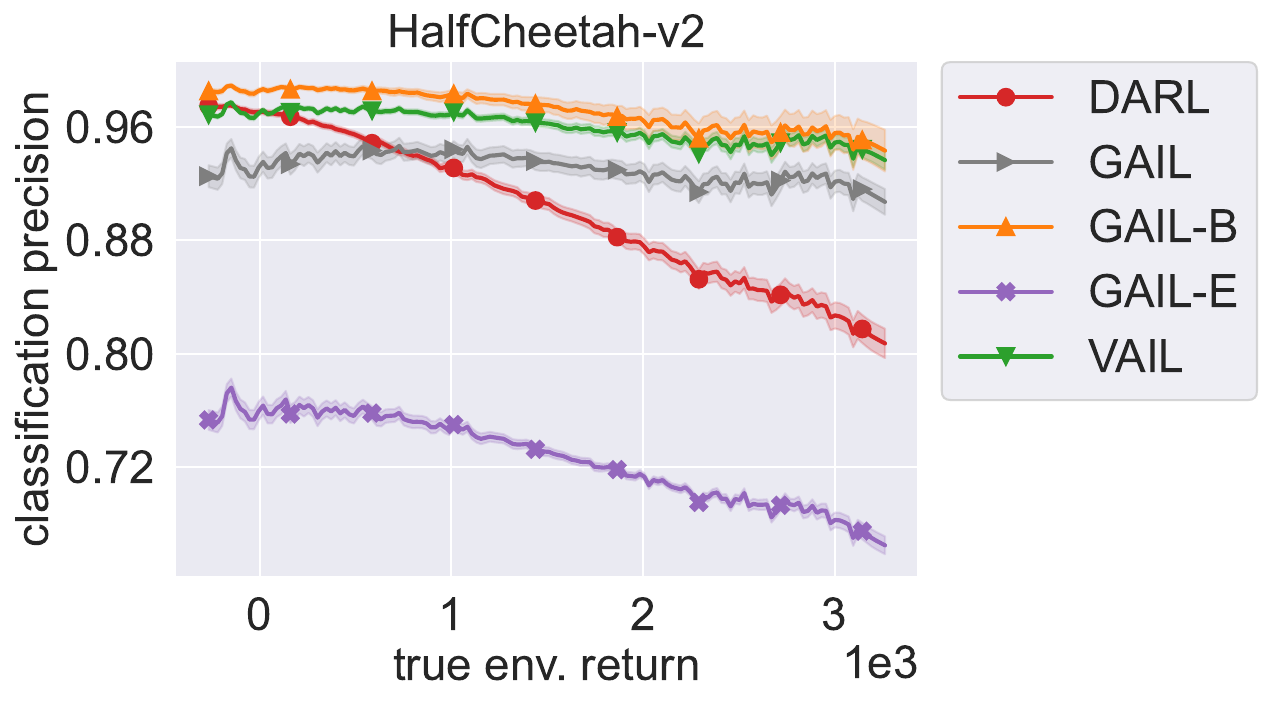}
\caption{Classification precision curves v.s. true environment return.} 
         \label{fig_exp_fix_policy_dependency_derived_reward}
     \end{subfigure}
        \caption{Derived return and classification precision of the discriminators or discriminator ensembles learned by various methods.}
        \label{fig_exp_fix_policy_denpendency_halfcheetah}
\end{figure}

After observing the policy dependency phenomena, we investigate whether discriminator ensembles and other GAIL variants can mitigate this issue. In \cref{fig_exp_fix_policy_denpendency_halfcheetah}, we present the normalized derived return and classification precision of the discriminators or discriminator ensembles trained by various baselines. The sample data is the same as that in \cref{sec_policy_dependency_problem}.
From \cref{fig_exp_fix_policy_dependency_classification_precision}, we observe that GAIL provides a relatively high derived return to data with a true return near $0$. As the true return increases, the derived return gradually decreases until the true return reaches $1,000$. After this point, the derived return begins to increase again. However, when the true return is between $2,500$ and $3,000$, the derived return decreases once more.
The reason for these abnormal drops in derived return, as shown in \cref{fig_exp_fix_policy_dependency_derived_reward}, is that the discriminator fails to classify low-return data accurately. The classification precision for $0$-return data is even lower than for data with a $1,000$ return. Ideally, $0$-return data should be more distinct from expert data than $1,000$-return data and thus easier to classify. This erratic trend in the derived return of GAIL can mislead policy learning, steering the policy away from optimal expert behaviors.

We also observe that the GAIL variants that have a single discriminator can alleviate policy dependency to different degrees. The derived-return drops at the beginning of the curves are alleviated. However, when the true return is between $2,500$ and $3,000$, the derived returns still decline. In contrast, \ours and GAIL-E, which employ a discriminator ensemble as the reward function, do not exhibit derived-return drops in this true return range. Furthermore, \ours shows the highest positive correlation with the true return. These results suggest that \ours can effectively alleviate the policy dependency problem and provide more accurate rewards to policies.

\begin{figure*}[ht]
    \centering
    \begin{subfigure}[b]{0.48\textwidth}
     \centering
     \includegraphics[width=\textwidth]{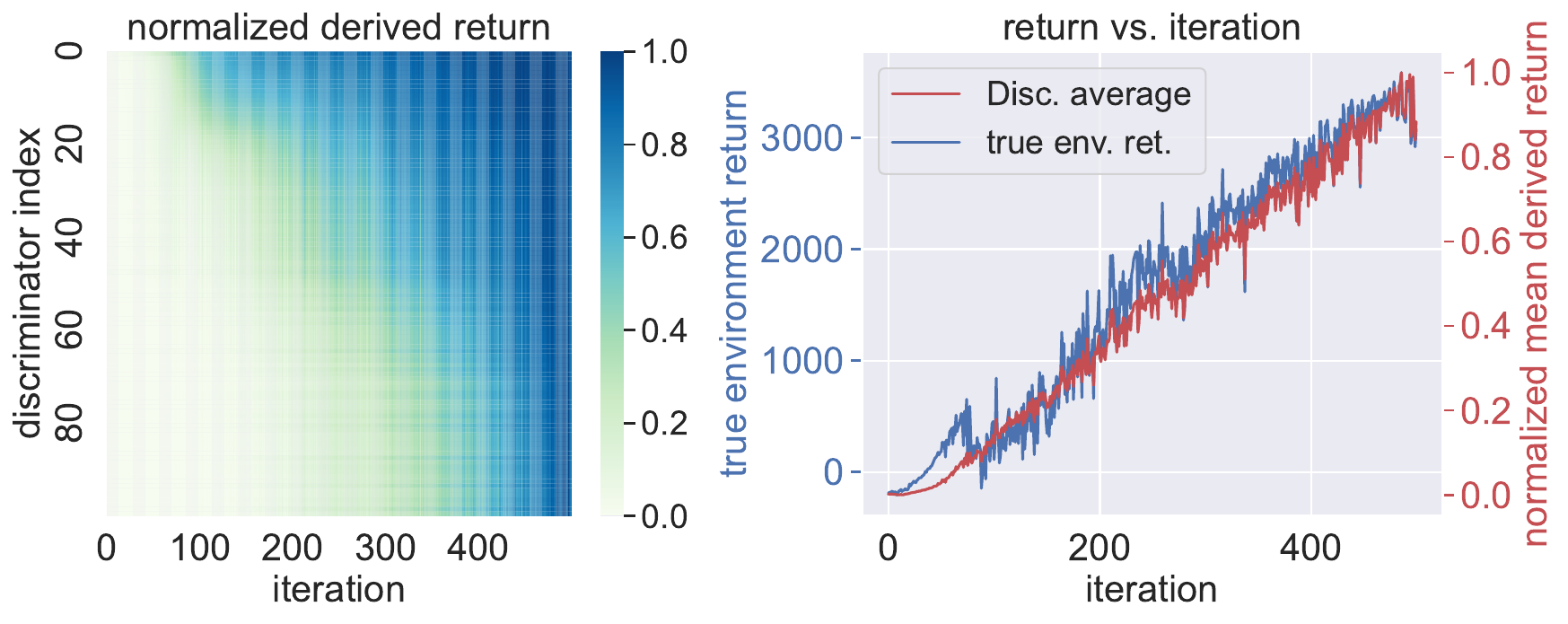}
     \caption{train the policy with the ensemble discriminator}
     \label{fig_training_process_grid_view_darl}
     \end{subfigure}
     \hfill
     \begin{subfigure}[b]{0.48\textwidth}
     \centering
     \includegraphics[width=\textwidth]{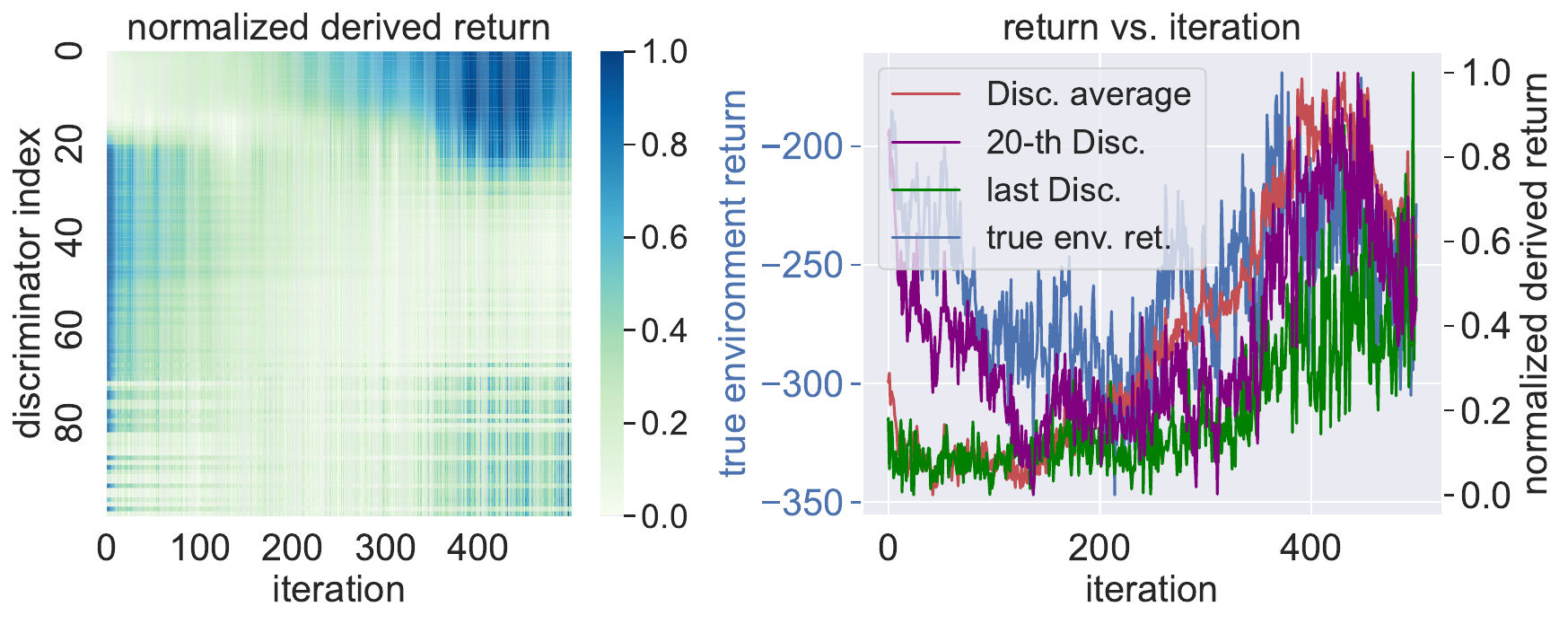}
     \caption{train the policy with the last discriminator}
     \label{fig_training_process_grid_view_gail}
     \end{subfigure}
    \caption{Normalized derived return and true environment return vs. policy training iteration in \texttt{HalfCheetah}. We train a policy using the ensemble discriminator (a) or the last discriminator (b) as the reward function. The left sub-figures in both panels depict the return derived by each discriminator for the policy at every iteration. Specifically, the point $(x,y)$, marked with color $c$, represents the normalized return derived by the $y$-th discriminator for the policy learned in the $x$-th iteration. The right sub-figures in both panels display the normalized returns derived by different reward models and the true environment returns vs. the policy training iterations. }
    \label{fig_training_process_grid_view}
\end{figure*}

To better understand how the ensemble of discriminators addresses the policy dependency problem, we analyze the behavior of each discriminator during policy training with the learned reward model. The policy is trained using the ensemble of learned discriminators as the reward function. During the training process, we record the return of the policy evaluated by each discriminator at every training iteration. The result is displayed in \cref{fig_training_process_grid_view_darl}.

We normalize the output of each discriminator to a range between $0$ and $1$. Here, $0$ denotes the lowest score for the policy in the current iteration throughout the training process for that discriminator, and $1$ denotes the highest score. The right sub-figure illustrates a steady increase in both the derived return and the true environment return. The left sub-figure reveals a clear trend: early-stage discriminators (those with smaller indices) quickly reach their maximum values in the initial training phases. As the policy's actual performance improves, the return output by later-stage discriminators begins to rise.

In this experiment, we see how the ensemble operates: in the early stages of training, the early discriminators are compatible with the current policy. The policy maximizes the output of these compatible discriminators, leading to an increase in the true environment return. As the actual performance of the policy improves, the index of compatible discriminators rises. Consequently, as the policy's performance continues to enhance, the output of the earliest discriminators gradually saturates and converges to the maximum value. Simultaneously, the output of the currently compatible discriminators begins to rise, further boosting the policy's performance. Thus, the discriminators in the ensemble operate in a relay-like manner, progressively improving the policy's performance.

In contrast, when we utilize only the last discriminator as the reward function to train the policy, the results are shown in \cref{fig_training_process_grid_view_gail}. The left sub-figure indicates that during training, the output of some discriminators (indices $\in[30, 60]$) gradually decreases, while the output of earlier discriminators initially increases before decreasing. To clarify these outputs, we have extracted the returns derived from the last discriminator, an early discriminator (the $20$-th discriminator), and an ensemble of discriminators, as illustrated in the right sub-figure. The output of the last discriminator increases gradually with training, whereas the true environment return oscillates—decreasing initially, then increasing, and subsequently decreasing again. Notably, the output trend of the early discriminator aligns more closely with the true environment return. This is because a poorly performing policy is better matched with the early discriminator, enabling it to more accurately reflect the policy’s actual performance. Furthermore, the mean output of all discriminators also follows a similar trend to the true environment return. This experiment highlights the policy-dependency problem: what the last discriminator regards as a good policy is considered incorrect by other discriminators, while the early discriminator, compatible with the current policy, can more accurately evaluate the policy's true performance.

\subsection{Alleviating Dynamics Dependency by \ours}
\label{sec_dynamics_dependency_solving}
\begin{figure}[ht]
 \centering
 \includegraphics[width=0.45\linewidth]{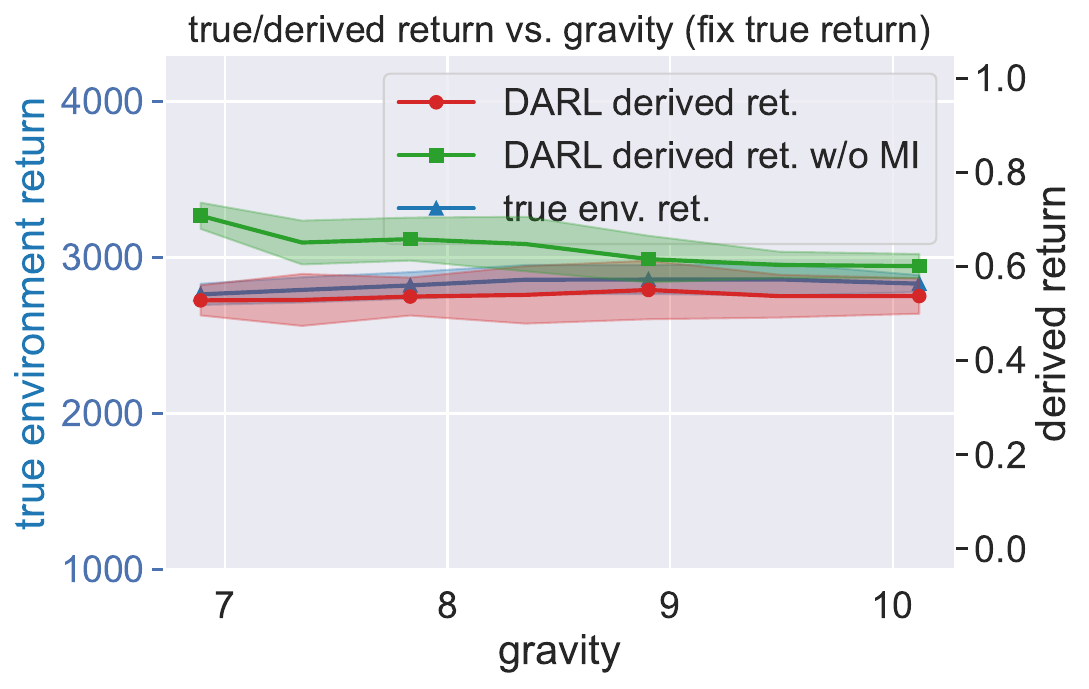}
\caption{True environment returns and derived returns vs. \texttt{gravity}. We trained a policy in \texttt{HalfCheetah} with \texttt{gravity} perturbations and sampled with this policy under different \texttt{gravity}. The figure is plotted using the trajectories with performance meeting $75\%$ expert return.}
     \label{fig_dynamics_gravity_curve_fix_ret}
\end{figure}

In this part, we investigate whether the \ours reward is dependent on the dynamics of the environment. We train a policy in \texttt{HalfCheetah} with \texttt{gravity} perturbations and sample trajectories using this policy with varying \texttt{gravity} dynamics. We select trajectories whose true environment return is approximately 75\% of the expert return, constructing a dataset with fixed true environment returns. We then use two types of \ours reward models to score these trajectories: one with the MI loss and one without it. \Cref{fig_dynamics_gravity_curve_fix_ret} illustrates the relationship between the derived return and \texttt{gravity} using the fixed-true-return dataset. We observe that when the true environment return remains constant with dynamics changes, the return derived from the reward model with the MI loss also does not vary with the dynamics. In contrast, without the MI loss, the return derived from the reward model decreases as \texttt{gravity} increases, indicating dynamics-dependent characteristics. This experiment demonstrates the dynamics dependency issue can be mitigated through MI minimization. We also confirm the dynamics-agnostics nature of the \ours rewards in \ref{supp_extened_dynamics_validata}.

\subsection{Reward Consistency}
\label{sec_reward_consistency}

\begin{figure}[ht]
\centering 
\includegraphics[width=0.89 \linewidth]{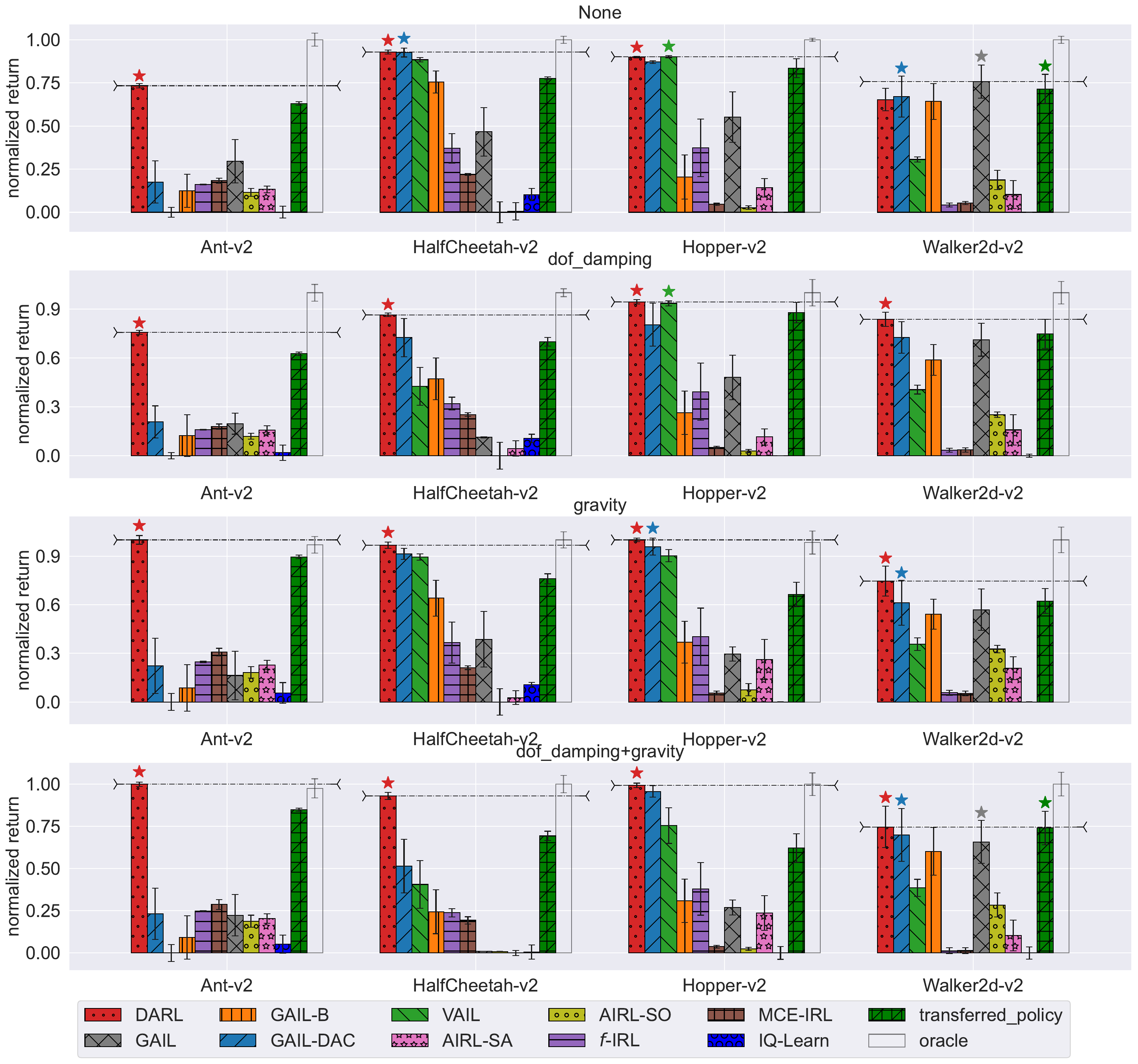}
\caption{Normalized final return of the policy trained with different reward functions in various transfer tasks. The methods significantly higher than other methods except for oracle are marked with \ding{72}. The changed dynamics parameters are in the title of each sub-figure.} 
\label{fig_main_comparison}  
\end{figure}

In this part, we aim to evaluate the alignment between the learned reward function $\hat{r}$ and the true reward function $r$. To achieve this, we introduce the concept of \textit{reward consistency}, a metric that assesses how accurately the learned reward reflects the true reward. A robust learned reward function $\hat{r}$ should ensure that if policy $\pi_1$ is preferred over $\pi_2$ under $\hat{r}$, the same preference should hold under the true reward $r$ for any pair of policies $(\pi_1, \pi_2)$ and any environment transition $p$. Formally, reward consistency is quantified as follows: (1) Train a policy for $B$ iterations using PPO with the learned reward function $\hat{r}$, denoting the policy at iteration $t$ as $\pi_t$; (2) Compute the reward consistency $RC(\hat{r})$ as the fraction of iterations where the direction of improvement under $\hat{r}$ aligns with that under the true reward, calculated by:
\begin{equation}
    \label{eq_reward_consistency}
    RC(\hat{r}) = {\sum_{t=1}^{B-1}\mathbb{I}\left\{\Delta(J^{\hat{r},p}, t)\cdot\Delta(J^{r,p}, t)>0\right\}}/{(B-1)},
\end{equation}
where $\Delta(J^{\hat{r},p},t) = J^{\hat{r},p}(\pi_{t+1}) - J^{\hat{r},p}(\pi_{t})$ represents the performance difference between consecutive iterations under the learned reward function, and $J^{r,p}(\cdot)$ and $J^{\hat{r},p}(\cdot)$ denote the expected returns estimated using the true reward $r$ and the learned reward $\hat{r}$, respectively. The metric $RC(\hat{r})$ effectively captures the proportion of policy updates that are correctly guided by the learned reward function.

We compare \ours with the GAIL variants in \texttt{gravity}-transfer tasks. The total policy training iteration $B$ is $1,000$. $J^{r,p}(\cdot)$ and $J^{\hat{r},p}(\cdot)$ are estimated by at least $10,000$ environment steps. The results are presented in \cref{tab_reward_consistency}. GAIL-B improves the reward consistency of GAIL in $3/5$ environments, which indicates that enlarging the training dataset can enhance the reward consistency. The reward consistency of GAIL-DAC, which additionally eliminates the reward bias by absorbing states, is no less than GAIL in all environments. These results imply that reducing overfitting (GAIL-B) and reward debiasing (GAIL-DAC) can mitigate policy dependency. Moreover, purely discriminator ensemble (GAIL-E) or mutual information minimization (VAIL) can only improve the reward consistency of GAIL in $3/5$ and $2/5$ environments, respectively. However, \ours, which minimizes mutual information and aggregates historical discriminators, achieves a significantly higher reward consistency compared to other GAIL variants. In all environments, \ours can guide the agent correctly with a probability of at least 92\%. This result directly demonstrates that \ours can provide more accurate rewards for policies in a transfer scenarios, with the reward consistency being significantly superior to the baseline methods.

\begin{figure}[ht]
\centering 
\includegraphics[width=0.9 \linewidth]{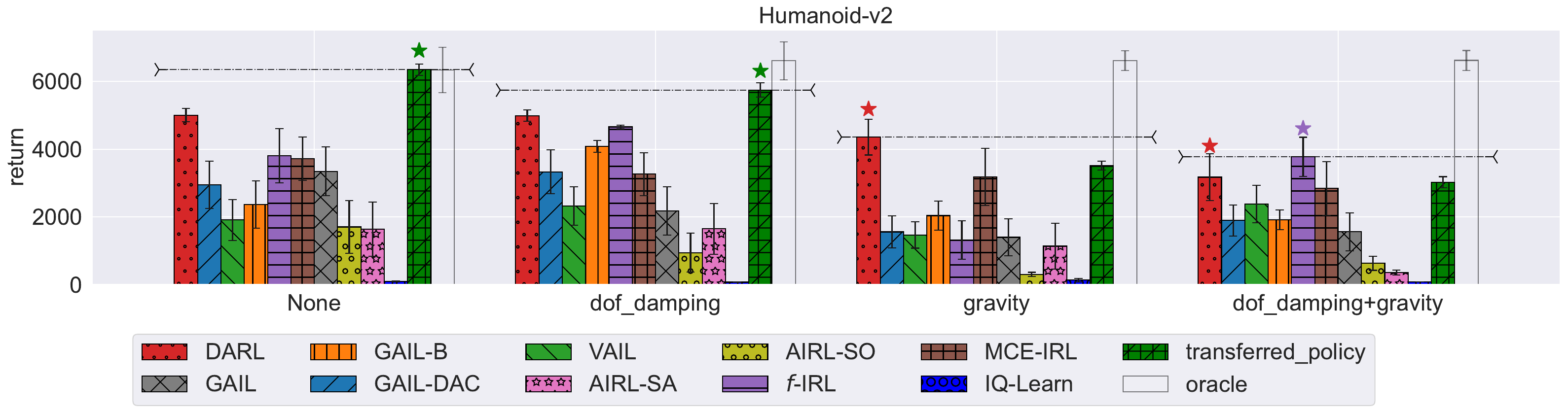}
\caption{Final return of the policy trained with different reward functions in \texttt{Humanoid}. The methods significantly higher than other methods except for oracle are marked with \ding{72}. The changed dynamics parameters are under the bottom of each sub-figure.} 
\label{fig_hard_task_humanoid}  
\end{figure}

\subsection{Performance in Various Transfer Tasks}
\label{sec_performance_comparison}

To investigate the impact of learned rewards on policy learning, we employ the reward from \cref{sec_reward_consistency} to train policies across various transfer tasks: no transfer (\texttt{None}), perturbations in \texttt{dof\_damping} or \texttt{gravity}, and both perturbations combined. Each policy is trained to convergence. Specifically, policies using PPO are trained with $1e7$ environment steps, while those utilizing SAC use $2e6$ steps. This discrepancy is due to the higher data efficiency of SAC and its increased computational time compared to PPO. These training durations have been empirically validated to ensure convergence.
The normalized return of each method in each task is presented in \cref{fig_main_comparison}. The method with the highest return in each task is normalized to $1.0$, and the one with the lowest return is set to $0.0$.

\Cref{fig_main_comparison} shows \ours obtains the highest return (except for oracle) in $15$ out of $16$ tasks. In the task where \ours fails to obtain the highest return, \ours is still close to the best baselines. Moreover, \ours gets the highest scores in all of the $12$ tasks with dynamics transfer, implying the strong robustness of \ours to dynamics transfer. \ours is significantly superior to \texttt{transferred\_policy} in $14/16$ tasks. Note that \texttt{transferred\_policy} means the transfer performance of the converged policy learned in the original environment by GAIL. The result means that the policy can adapt to the new environment rather than simply mimicking the expert behaviours. If the reward function only guided policies to replicate expert behavior precisely, surpassing the \texttt{transferred\_policy} would be difficult. Conversely, other AIL/IRL baselines rarely exceed the \texttt{transferred\_policy}. In transfer tasks, only GAIL-DAC and VAIL occasionally surpass the \texttt{transferred\_policy}. Achieving a policy superior to direct transfer is significant as it demonstrates the potential for policy improvement through reward transfer, making policy training with the learned reward non-trivial. Besides, the improvement of \ours over other baselines is the most remarkable in \texttt{Ant} and \texttt{HalfCheetah}, especially in \texttt{dof\_damping+gravity} transfer tasks. \texttt{HalfCheetah} is an environment without a terminal signal, where the trajectory length is fixed. In \texttt{HalfCheetah}, the agent only needs to maximize the mean reward rather than the trajectory length. Therefore, the training efficiency in \texttt{HalfCheetah} directly relate to the reward correctness and consistency. The remarkable performance improvement of \ours in \texttt{HalfCheetah} further underscores its high reward accuracy and consistency. \texttt{Ant} is a relatively hard environment as the dimension of its state and action is the largest among the $4$ environments in \cref{fig_main_comparison}. The substantial improvement of \ours in \texttt{Ant} indicates its scalability to more complex environments.
\begin{figure}[ht]
\centering 
\includegraphics[width=0.7 \linewidth]{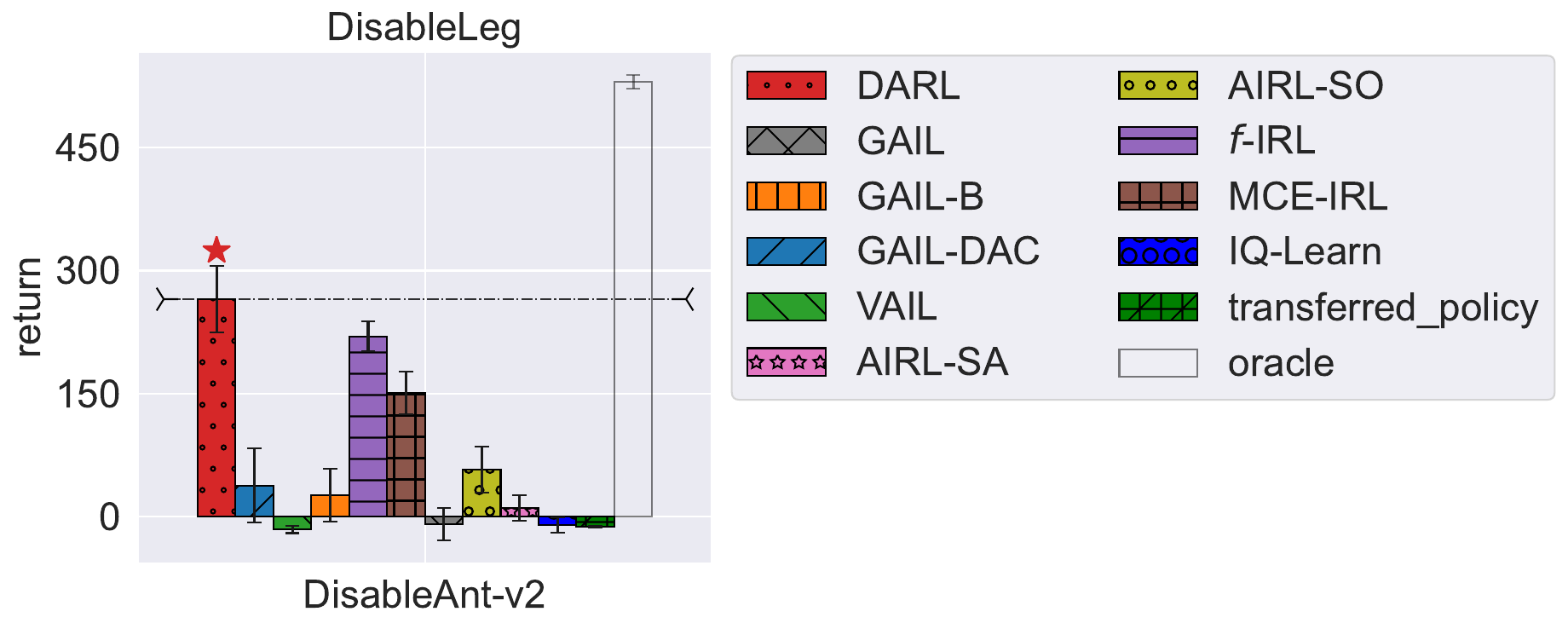}
\caption{Final return of the policy trained with different reward functions in \texttt{DisableAnt}. The methods significantly higher than other methods except for oracle are marked with \ding{72}.} 
\label{fig_hard_task}  
\end{figure}

\begin{figure*}[htbp]
\centering 
\includegraphics[width=1.0 \linewidth]{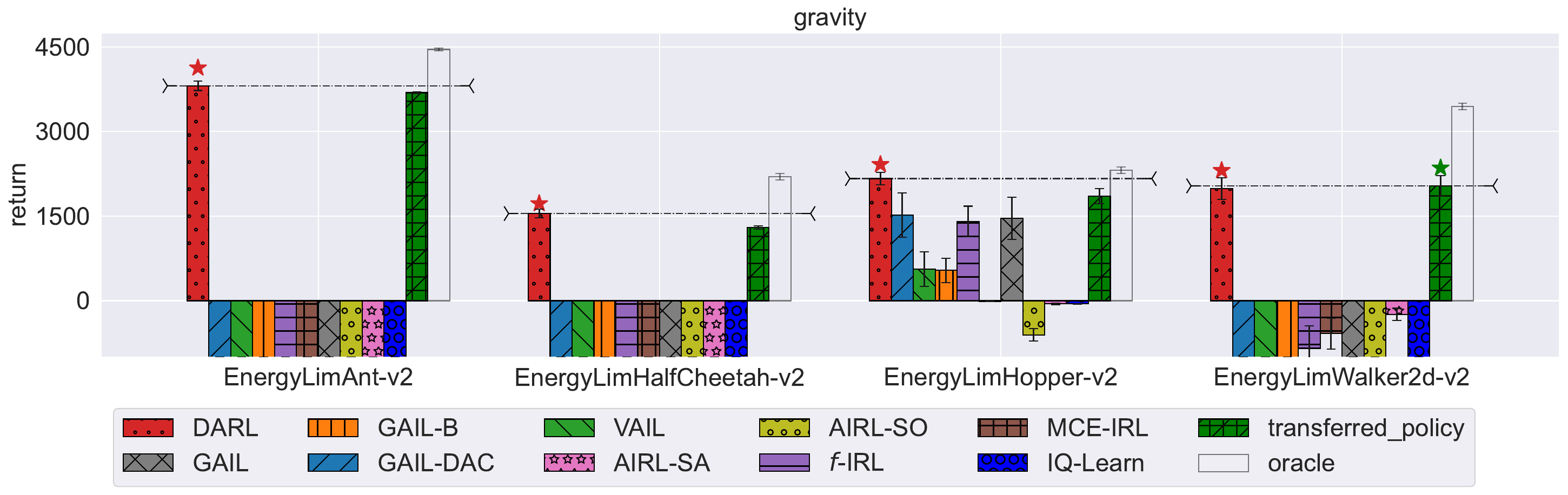}
\caption{Final return of the policy trained with different reward functions in energy-limit transfer tasks. The methods significantly higher than other methods except for oracle are marked with \ding{72}.  We clip the value less than $-1,000$. We change \texttt{gravity} in these environments.} 
\label{fig_rampti_task}  
\end{figure*}

\begin{figure*}[ht]
     \centering
     \begin{subfigure}[b]{0.47\linewidth}
         \centering
         \includegraphics[width=1.0 \linewidth]{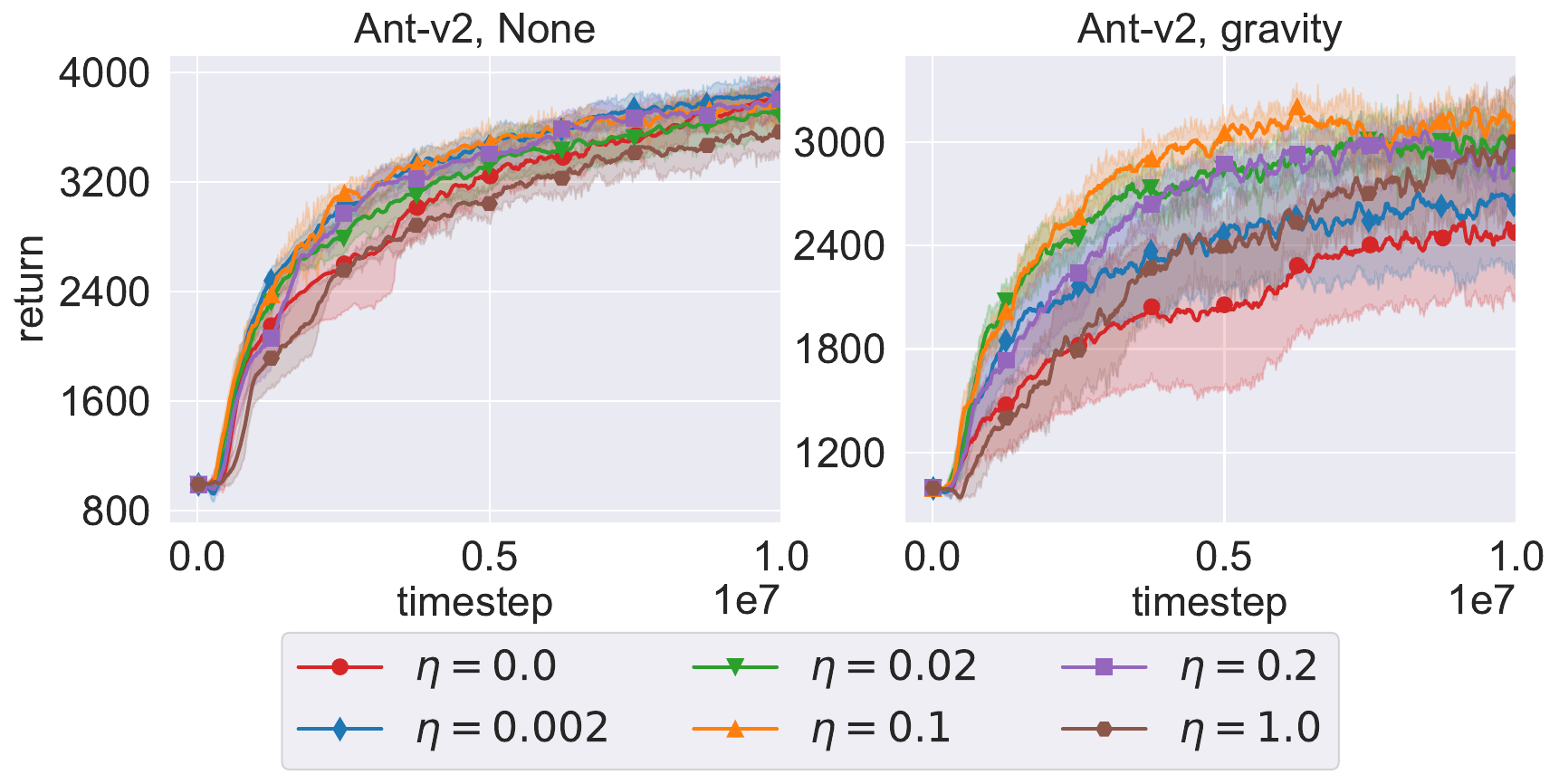}
\caption{Ablation studies on the regularization factor $\eta$.} 
         \label{fig_ablation_on_eta}
     \end{subfigure}
     % \hfill
     \begin{subfigure}[b]{0.47\linewidth}
         \centering
         \includegraphics[width=1.0 \linewidth]{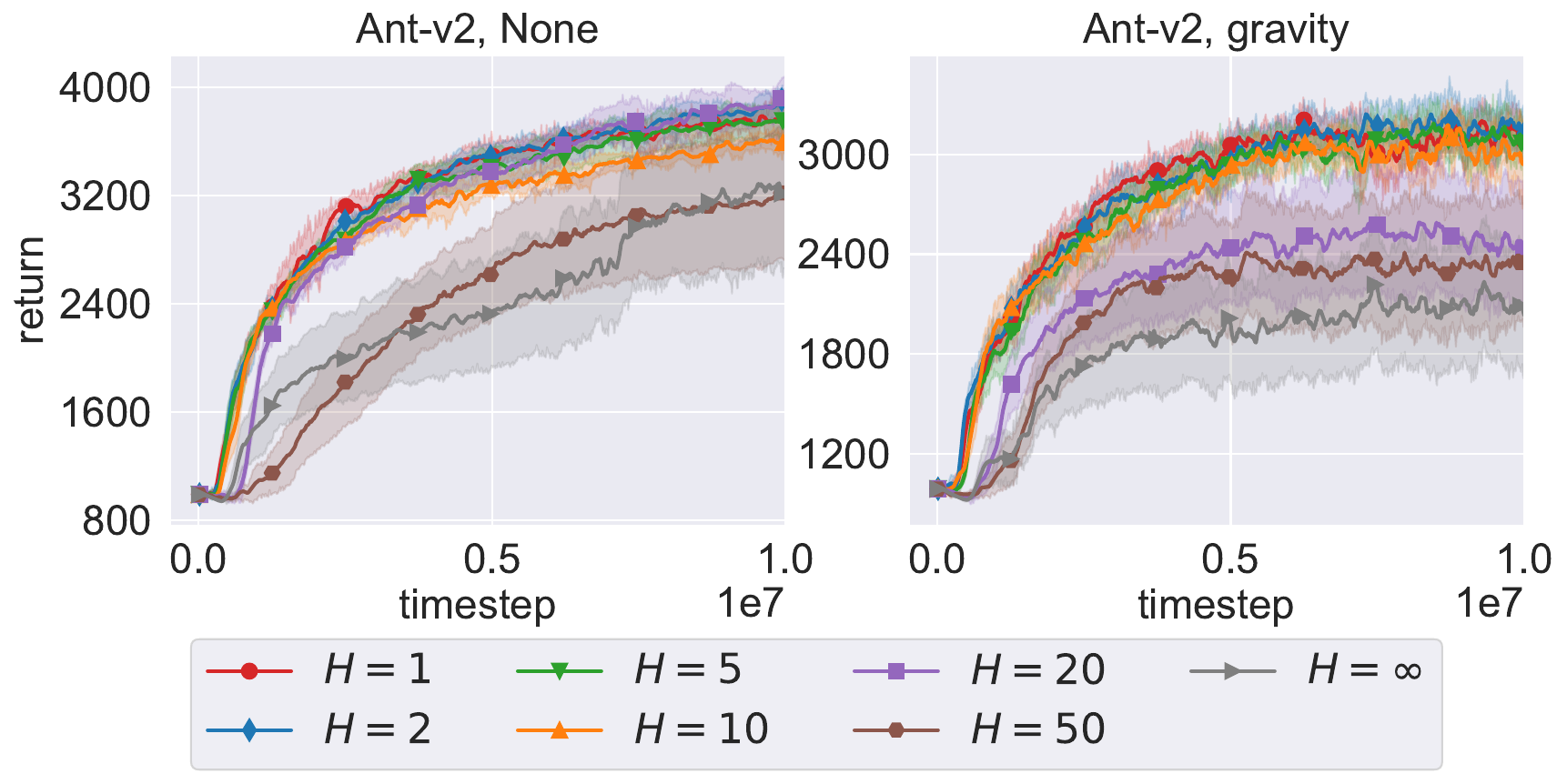}
\caption{Ablation studies on the backup interval $H$.} 
         \label{fig_ablation_on_t}
     \end{subfigure}
        \caption{Learning curves shaded with one standard error in original and transfer tasks with different regularization factor $\eta$ and backup interval $H$. $H=\infty$: the ensemble technique is not adopted and the discriminator at the convergence is used as the reward. }
        \label{fig_ablation}
\end{figure*}

\subsection{Performance in More Difficult Tasks}
\label{sec_performance_comparison_difficult}

We conduct the experiments in \texttt{DisableAnt} and \texttt{Humanoid} to further assess the scalability of \ours for more difficult tasks. In \texttt{DisableAnt}, survival in the transfer environment requires the agent to behave significantly differently from the expert in the original environment~\cite{fu2017airl}. In \texttt{Humanoid}, the dimensions of states and actions are $376$ and $17$, respectively, which are far larger that the ones of \texttt{Ant}. We first train the reward function in the original \texttt{Ant} and \texttt{Humanoid} without transfer. Then, we train the policy with the learned reward function in \texttt{DisableAnt} and dynamics-transfer \texttt{Humanoid}. The results in \texttt{DisableAnt} are presented in \cref{fig_hard_task}. Previous works~\cite{ni2020firl,fu2017airl,zeng2022mlirl} show that state-only reward functions work better in \texttt{DisableAnt}. Therefore, we learn a state-only reward function for MCE-IRL in the environment. In \texttt{DisableAnt}, the \texttt{transferred\_policy} completely fails to transfer, indicating the large gap between \texttt{Ant} and \texttt{DisableAnt}. Moreover, except for \ours and oracle, the highest returns are obtained by $f$-IRL, MCE-IRL, and AIRL-SO, all of which learn state-only reward functions. This finding aligns with previous studies~\cite{ni2020firl,fu2017airl}, suggesting that state-only reward functions help decouple from dynamics. However, \ours, which learns state-action reward function, demonstrates better transfer performance. This result reveals that a transferable reward does not necessarily need to be state-only but can depend on both state and action simultaneously, which can even perform better. 

The results of \texttt{Humanoid} are presented in \cref{fig_hard_task_humanoid}, demonstrating that \ours achieves or exceeds the performance of existing AIL/IRL algorithms in all four scenarios. This further validates \ours's robustness in high-dimensional input-output environments. However, in scenarios where the environment dynamics are unchanged or only \texttt{dof\_damping} is changed, DARL does not outperform the \texttt{transferred\_policy}. Additionally, in the case where both \texttt{dof\_damping} and \texttt{gravity} are changed, \ours surpasses \texttt{transferred}\_\texttt{policy} but does not perform as well as $f$-IRL. By analyzing \ours's reward consistency across these four tasks, as shown in \cref{tab_consistency_humanoid}, we found that \ours consistently exhibits high reward consistency (greater than $0.92$). However, the final policy performance in \cref{fig_hard_task_humanoid} does not show a significant advantage as observed in other environments in \cref{fig_main_comparison}. This discrepancy may be attributed to the limitations of reinforcement learning algorithms in high-dimensional input-output scenarios, where PPO struggles to perform robust training in complex environments with noisy rewards, thereby limiting policy performance improvements. This phenomenon highlights a limitation of \ours: it cannot completely and accurately restore the true environment rewards, which may generate reward noise and impact the final policy performance in large-scale environments.

\begin{table}[ht]
    \centering
    \caption{Reward consistency $\pm$ standard error in \texttt{Humanoid} with different dynamics changes.}
    \label{tab_consistency_humanoid}
    \begin{tabular}{r@{~$\pm$~}lr@{~$\pm$~}lr@{~$\pm$~}lr@{~$\pm$~}l}\toprule
\multicolumn{2}{c}{None} & \multicolumn{2}{c}{dof\_damping}  & \multicolumn{2}{c}{gravity} & \multicolumn{2}{c}{dof\_damping  \& gravity}\\\midrule
$ 0.92 $ & $ 0.01 $ & $ 0.93 $ & $ 0.00 $ & $ 0.92 $ & $ 0.01 $& $ 0.92 $ & $ 0.00 $ \\
\bottomrule
    \end{tabular}
\end{table}

\subsection{Performance in Action-Dependent Tasks}
\label{supp_action_dependent}

\ours aims to learn a state-action reward function as there exist a tremendous amount of real-world tasks whose reward functions are action-dependent. To demonstrate \ours can also handle tasks with reward functions dependent on both state and action, we modify the reward function of MuJoCo tasks. The major parts of the reward function of a MuJoCo task are (i) forward velocity, (ii) alive bonus, and (iii) action cost. For instance, in \texttt{Hopper}, the reward function is defined as:
\begin{equation}
\label{eq_reward_of_hopper}
r = \underbrace{\left(x' - x\right)/\text{dt}}_{\text{forward velocity}} - 0.001\underbrace{\left\Vert \mathbf{a} \right\Vert_2^2}_{\text{action cost}} + \underbrace{1.0}_{\text{alive bonus}},
\end{equation}
where $x$ and $x'$ represent the robot's position at the current and next time steps, respectively, $\mathbf{a}$ denotes the action, and $\text{dt}$ is the control period. The term $\left(x' - x\right)/\text{dt}$ approximates the velocity through the position difference. Since the robot's velocity can be derived from the state, the first part of the reward can be inferred directly from the state. Thus, only the second part, the action cost, is action-related. However, the coefficients of the action cost in MuJoCo tasks are always small, resulting in the reward being dominated by the state-related and constant components. Therefore, the reward functions of MuJoCo tasks can be approximated as being primarily state-dependent.

We now consider a class of energy-limited locomotion tasks, where the objective is to control a robot to run forward quickly while minimizing energy consumption. Let the original reward function be denoted as $r$. In energy-limited environments, the reward function is modified as follows:
\begin{equation}
\label{eq_modified_reward_of_hopper}
r_\text{energy-limit} = r - 3\left\Vert \mathbf{a} \right\Vert_2^2.
\end{equation}
\eqref{eq_modified_reward_of_hopper} increases the penalty coefficient by a factor of 3,000, resulting in a reward that is heavily dependent on the action.
\begin{table}[htbp]
\centering
\caption{Average return of the expert trained using the default MuJoCo reward function. The expert demonstrations are evaluated using both the default reward function and the energy-limited reward function.}
\begin{tabular}{l|cccc}\toprule
   Rew. func.        & Ant-v2 & HalfCheetah-v2 & Hopper-v2 & Walker2d-v2 \\\midrule
Default    & $2958.42$   & $3930.16$           & $3315.92$      & $3272.10$     \\
Energy-limit & $1259.93$   & $-3730.90$          & $1937.10$      & $1176.58$    \\\bottomrule
\end{tabular}
\label{tab_new_reward_expert_score}
\end{table}
We use the reward in \eqref{eq_modified_reward_of_hopper} to evaluate the demonstrations in \cref{sec_performance_comparison}, which are collected by the expert policy trained using the default MuJoCo reward function. The results are presented in \cref{tab_new_reward_expert_score}. It is observed that the action cost significantly impacts the expert reward, indicating that the action-related reward becomes a substantial part of the total reward. Policies that overlook action cost fail to achieve high returns, particularly in \texttt{HalfCheetah}.

After training expert policies with the energy-limit reward and collecting expert demonstrations, we learn the reward functions and train policies using the learned reward function in energy-limited environments with \texttt{gravity}-transfer. Results are demonstrated in \cref{fig_rampti_task}, where tasks prefixed by \texttt{EnergyLim} represent the energy-limit environments. Among the various AIL and IRL methods evaluated, only \ours achieved positive returns in \texttt{EnergyLimAnt}, \texttt{EnergyLimHalfCheetah}, and \texttt{EnergyLimWalker2d}. This indicates that other methods may not effectively correlate actions with the corresponding reward functions, which is a particularly notable issue in state-only approaches. Such methods might mistakenly prioritize rapid movement over energy conservation, resulting in negative returns despite achieving quicker movement speeds. In contrast, \ours excelled in conserving energy, as evidenced by its performance, which significantly exceeded the baseline policy in three out of four scenarios. These results imply that  \ours effectively utilizes action information to infer state-action dependent reward functions. \footnote{The \texttt{EnergyHumanoid} task was excluded due to difficulties in obtaining a satisfactory expert policy using PPO or SAC.}

\subsection{Ablation Studies}
\label{sec_ablations}

In this part, we investigate the impact of the regularization factor $\eta$ in \eqref{eq_modified_d_loss}, the discriminator backup interval $H$, and the reward transformation techniques described in \eqref{eq_reward_transform_trick_normalize} and \eqref{eq_reward_transform_trick_clip}. Our experiments are conducted using the \texttt{Ant} environment, both without transfer and with \texttt{gravity}-transfer.

\noindent{\textbf{Regularization factor} $\mathbf{\eta}$.}
First, we fix $H$ to 1 and vary $\eta$. The results are presented in \cref{fig_ablation_on_eta}. In the no-transfer task, all values of $\eta$ converge to a similar final return, except for the maximum $\eta$ (1.0), which is inferior to others by a noticeable margin. This indicates that an excessively large $\eta$ can cause the discriminator to discard essential information from its input, thereby degrading policy performance. In contrast, the returns for different $\eta$ vary significantly in the \texttt{gravity}-transfer task (the right figure in \cref{fig_ablation_on_eta}). As $\eta$ increases from $0.0$ to $0.1$, the dynamics-related information used by the discriminator gradually decreases, and the return correspondingly increases. This indicates that the reward functions using less dynamics-related information will be more robust to the dynamics transfer. However, as $\eta$ increases further from $0.1$ to $1.0$, the return decreases due to the excessive elimination of information. Despite this, larger $\eta$ values still demonstrate better robustness to dynamics transfer compared to smaller $\eta$ values ($0.0$ or $0.002$).

\begin{figure}[htbp]
\centering 
\includegraphics[width=0.7 \linewidth]{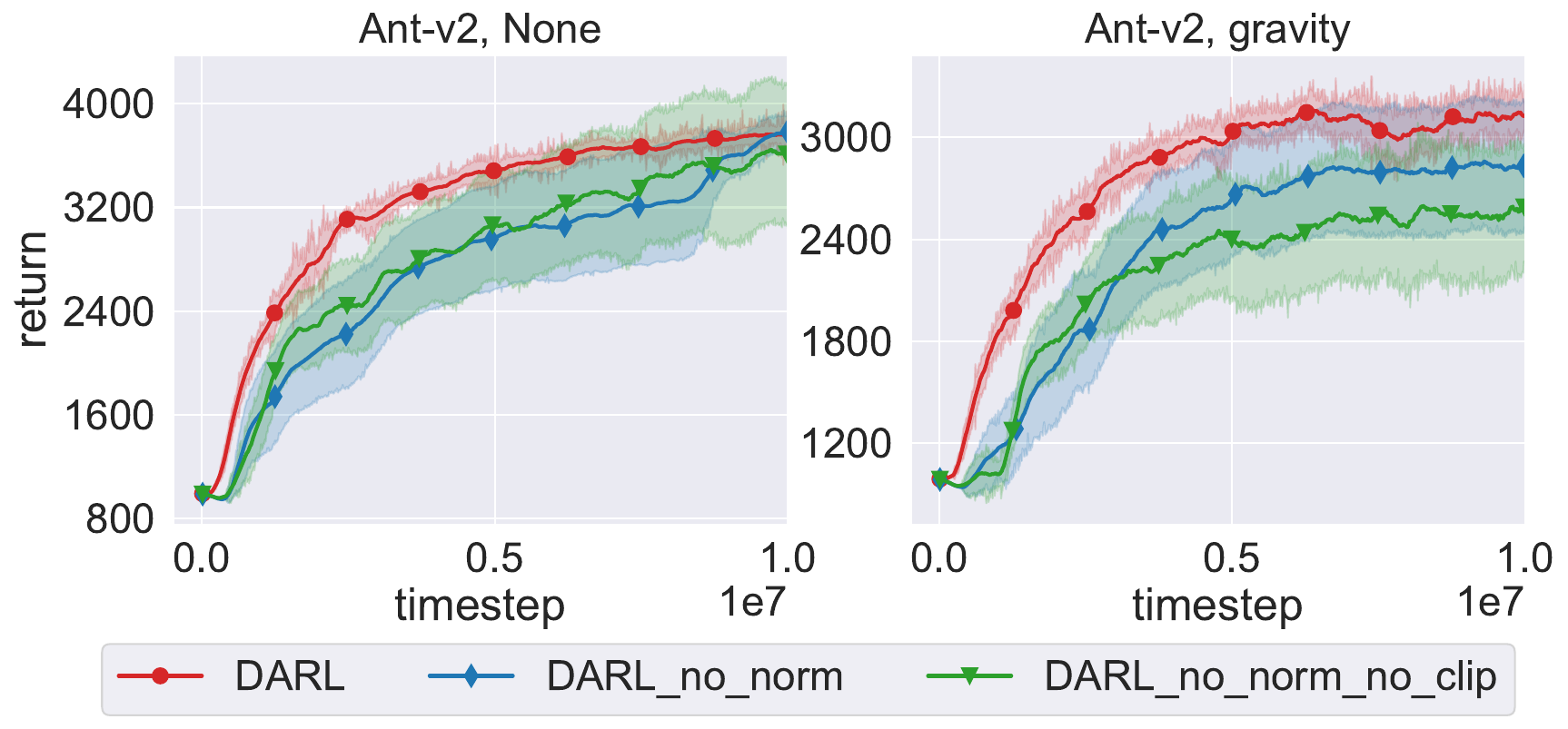}
\caption{Learning curves shaded with one standard error in original and transfer tasks with different reward normalization or clipping tricks.} 
\label{fig_ablation_reward_transform}  
\end{figure}

\noindent{\textbf{Discriminator backup interval} $H$.}
In \cref{fig_ablation_on_t}, we fix $\eta$ at $0.1$ and vary $H$. The results from both the no-transfer and \texttt{gravity}-transfer tasks indicate that smaller values of $H$ yield higher returns. With a total training iteration of $100$, the largest backup intervals ($H=\infty$ and $H=50$) correspond to the smallest ensemble sizes, specifically $1$ and $2$, respectively. These small ensemble sizes result in the lowest returns for both tasks. Although dynamics-related information is properly eliminated, a small number of discriminators struggle to provide accurate policy guidance. As $H$ is reduced to $20$, returns improve in the no-transfer task but remain poor in the \texttt{gravity}-transfer task. Conversely, smaller $H$ values lead to higher returns in both tasks, indicating that larger ensemble sizes enhance the reward robustness in dynamics transfer.

We also observe that a larger discriminator ensemble consumes more memory. In \ref{supp_memory_cost}, we analyze the memory and time costs associated with the discriminator ensemble, finding them to be tolerable. The largest discriminator of \ours in our experiments is only $0.428$MB. As we reserve $100$ discriminators to construct the ensemble, there is also $42.8$MB of additional memory cost for the discriminator ensemble. We believe that the performance improvements shown in \cref{fig_ablation_on_t} justify the memory cost.

In conclusion, a robust reward function with high reward consistency requires an appropriately chosen $\eta$ and a sufficiently large ensemble size. A too small $\eta$ can injure the robustness of dynamics transfer, while too large $\eta$ may eliminate essential information, leading to a performance drop. Additionally, a too small ensemble size may fail to achieve high reward consistency and is less robust to dynamics transfer.

\noindent{\textbf{Reward normalization and clip tricks}.} When aggregating the outputs of the discriminator ensemble, we normalize each discriminator's output and clip the maximum output to a constant, as shown in \eqref{eq_reward_transform_trick_normalize} and \eqref{eq_reward_transform_trick_clip}. We conduct ablation studies on these normalization and clipping tricks to evaluate their effects. The results, depicted in \cref{fig_ablation_reward_transform}, indicate that the final returns of the three variants in the no-transfer task are comparable. However, the learning curve of \ours is smoother with reduced shading, suggesting that these techniques stabilize the training process. In the \texttt{gravity}-transfer task, the asymptotic performance of the variants differs, with more tricks leading to higher performance. \Cref{fig_ablation_reward_transform} confirms that the reward transformation techniques enhance both training stability and asymptotic performance. Further details on the normalization technique are provided in \ref{app_influence_of_changing}.

\begin{figure}[ht]
    \centering
    \includegraphics[width=0.7\linewidth]{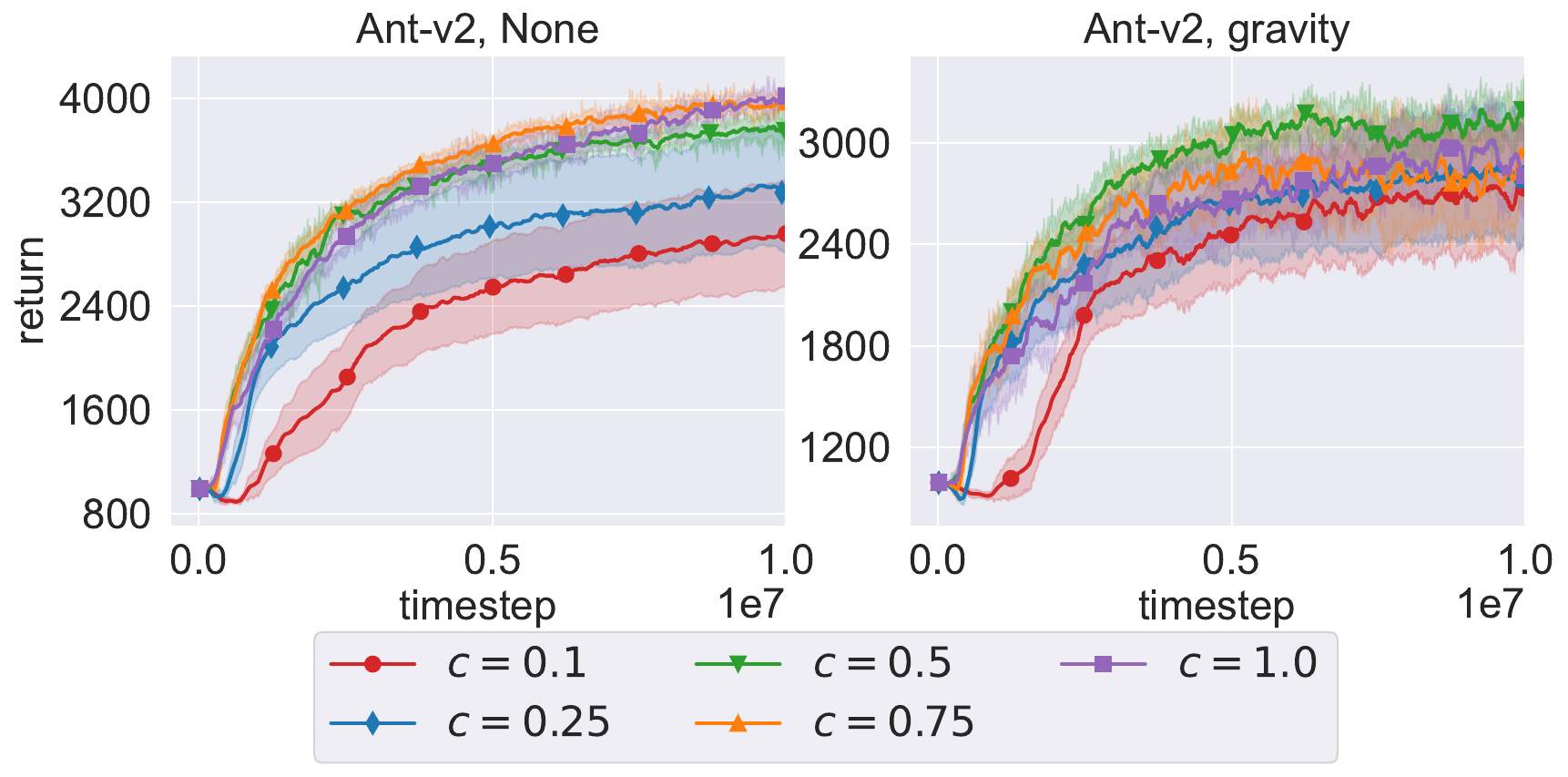}
    \caption{Learning curves shaded with one standard error in original and transfer tasks with different clipping factor $c$.}
    \label{fig_ablation_c_ant}
\end{figure}

\noindent{\textbf{Clipping ratio} $c$.} To analyze the impact of the clipping ratio $c$ in \eqref{eq_reward_transform_trick_clip} on training, we fix $\eta=0.1$ and $T=1$ and vary $c$ as shown in \cref{fig_ablation_c_ant}. In environments without dynamic perturbations, lower values of $c$ ($0.1$ and $0.25$) significantly reduce policy performance due to excessive suppression of discriminator output. Higher values of $c$ exhibit less notable performance differences. In environments with dynamics transfer, the performance at $c=0.5$ surpasses other settings, indicating that both very high and very low values of $c$ hinder training. A balanced value of $c=0.5$ yields optimal results across all environments, leading us to standardize $c$ at this value.

\subsection{Embedding Visualization}

\begin{figure}[ht]
 \centering
 \begin{subfigure}[b]{0.45\linewidth}
     \centering
     \includegraphics[width=\textwidth]{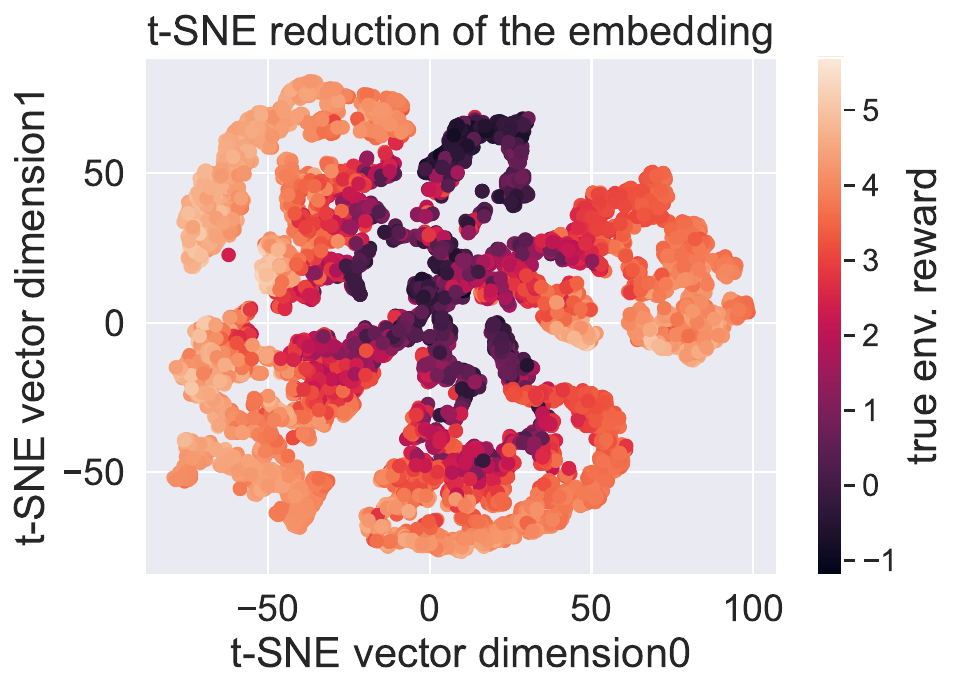}
     \caption{with the MI loss.}
     \label{fig_response_tsne_with_mi_loss}
 \end{subfigure}
 % \hfill
 \begin{subfigure}[b]{0.45\linewidth}
     \centering
     \includegraphics[width=\textwidth]{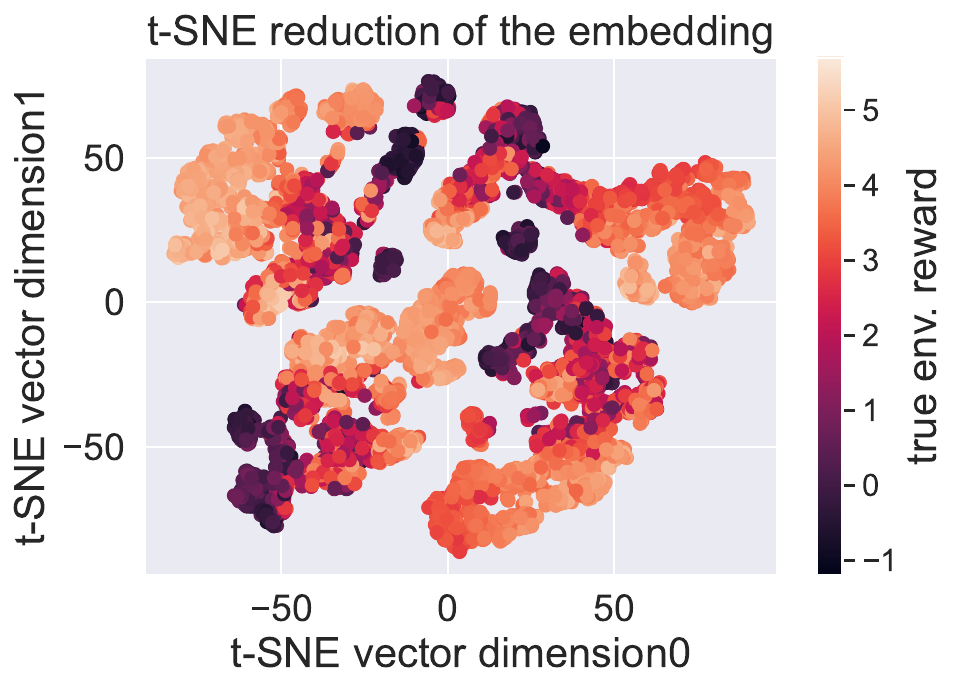}
     \caption{without the MI loss.}
     \label{fig_response_tsne_without_mi_loss}
 \end{subfigure}
    \caption{t-SNE visualizations of the encoder network $E(s,a;\phi)$ outputs, with and without the utilization of mutual information (MI) loss as in \eqref{eq_loss_encoder}. The encoder outputs are mapped to a 2-dimensional space using t-SNE. The $x$- and $y$-axes represent the first and second dimensions of the reduced-dimensionality vectors, respectively, and the point colors indicate the corresponding true environment rewards. The data is sampled from the \texttt{HalfCheetah} environment.}
    \label{fig_response_tsne}
\end{figure}
To discover what the encoder has learned, we collected a batch of data generated during the training process. We then mapped the state and action to the embedding space and used t-SNE~\cite{van2008visualizing} for dimensionality reduction, , projecting the embeddings into two dimensions as shown in \cref{fig_response_tsne}. In this figure, colors represent the true environment rewards corresponding to each embedding. We observe that with the mutual information loss, the embeddings strongly correlate with the true environment reward—the closer an embedding is to the center, the higher the true reward. Conversely, without the mutual information loss, this relationship vanishes, suggesting that the embeddings retain a significant amount of information unrelated to the reward.

\section{Conclusions and Future Work}
\label{sec_conclusion}
\noindent{\textbf{Conclusions.}} In this work, we introduced \ours, an IRL method based on the AIL framework, designed to derive transferable state-action reward functions robust to dynamics transfer. To reduce the dependency of discriminators on dynamics and behavior policies, we incorporated a mutual information minimization loss and proposed a discriminator ensemble method. Empirical evaluations across five MuJoCo tasks with both state-only and state-action rewards demonstrate that \ours can learn reward functions more consistent with the true reward than previous methods, leading to policies with higher returns in transferred scenarios. % This work contributes to the AI community by enhancing the robustness and transferability of reward functions in reinforcement learning. This advancement can lead to more adaptable and reliable AI systems in dynamic and variable environments.

\noindent{\textbf{Limitations and future work.}} In environments with high-dimensional state spaces, such as those with image inputs, \ours may struggle to satisfy \eqref{eq_club_constraint} due to the challenge of approximating $p(s'| z)$. A potential future direction is to introduce an encoder to map the high-dimensional state to a low-dimensional embedding space. In this paper, we made minimal modifications to GAIL, just adding an auxiliary loss and a discriminator ensemble to develop \ours. Another promising future direction is to integrate \ours with other reward debiasing approaches like DAC~\cite{kostrikov2019dac} to eliminate reward biases, particularly in finite-horizon MDPs where the existence of the absorbing state is often overlooked~\cite{sutton1998rl}. A third potential direction is to reduce memory costs. Although the memory cost of the discriminator ensemble was tolerable in our experiments (\cref{sec_ablations}), it increases more rapidly than that of a single discriminator as model complexity grows. Substituting the discriminator ensemble with last-iterate convergence no-regret learning methods, which do not require retaining historical discriminators and offer theoretical guarantees of convergence~\cite{golowich2020tight_last_iterate,lei2021last_iterate}, could be a valuable future work.

\bibliographystyle{unsrt}
\bibliography{darl_bib.bib}
\clearpage

% \newpage

%%%%%%%%%%%%%%%%%%%%%%%%%%%%%%%%%%%%%%%%%%%%%%%%%%%%%%%%%%%%

% \clearpage
\appendix
\section{Proof}

\subsection{Proof of Theorem~\ref{thm_emsemble}}
\label{proof_thm_emsemble}
\begin{theorem}[Discriminator Ensemble Upper Bound]
Consider finite $\mathcal{S}$ and $\mathcal{A}$, if $\max_{\pi\in\{\pi_t| t\in[1,T]\}}\| \nabla_D \mathcal{L}(D, \pi_t) \|_2 \leq G$, Alg.~\ref{alg_OGD} with step sizes $\{ \omega_t = \frac{\Omega}{G\sqrt{t}}, t=1,2,\cdots,T \}$ will achieve the following guarantee for all $T \geq 1$
\begin{equation}
    \frac{1}{T}\sum_{t=1}^T\mathcal{L}(\overline{D},\pi_t)  \leq \frac{1}{T}\min_D\sum_{t=1}^T \mathcal{L}(D,\pi_{t}) + \frac{3\overline{\Omega} G}{2\sqrt{T}},
\end{equation}
where $\Omega$ is a hyperparameter and $\overline{\Omega}=\max\{\frac{\|D_T-D^\star\|_2^2}{\Omega}, \Omega\}$.
\end{theorem}

\begin{proof}
Recall the definition in \eqref{eq_gail_loss}
\begin{align}
    \mathcal{L}(D, \pi) &= \lambda \mathcal{H}(\pi)-\mathbb{E}_{s,a\sim\mathcal{B}^E}[\log (D(s,a))] \nonumber\\
    &~~~~~~~~~~~~~~~~-\mathbb{E}_{s,a\sim\mathcal{B}^\pi}[\log(1-D(s,a))] .
\end{align}
Here each entry of $D\in(0,1)^{|\mathcal{S}|\times|\mathcal{A}|}$ corresponds to each state-action pair $(s,a)\in\mathcal{S}\times\mathcal{S}$.  
$\forall s\in\mathcal{S},a\in\mathcal{A}$, the gradient for $D(s,a)$ of $\mathcal{L}(D, \pi_t)$ is
\[
\begin{aligned}
    \nabla_{D(s,a)} \mathcal{L}(D, \pi_t) =& - \mathbb{E}_{s',a'\sim\mathcal{B}^E}\left[\frac{\mathbb{I}\{s'=s,a'=a\}}{D(s,a)}\right] +\\&~~~~~\mathbb{E}_{s',a'\sim\mathcal{B}^{\pi_t}}\left[ \frac{\mathbb{I}\{s'=s,a'=a\}}{1-D(s,a)} \right], \nonumber
\end{aligned}
\]
where $\mathbb{I}\{s'=s,a'=a\}$ is an indicator function, which returns $1$ if $s'=s,a'=a$ and $0$ otherwise. Then we can get the second-order gradient:
\begin{equation}
\label{A.1nabla1}
\begin{aligned}
    \nabla^2_{D(s,a)} \mathcal{L}(D, \pi_t) =& \mathbb{E}_{s,a\sim\mathcal{B}^E}\left[\frac{\mathbb{I}\{s'=s,a'=a\}}{D^2(s,a)} \right] + \\
    &\mathbb{E}_{s,a\sim\mathcal{B}^{\pi_t}}\left[ \frac{\mathbb{I}\{s'=s,a'=a\}}{(1-D(s,a))^2} \right] \\
    \geq &  0.
\end{aligned}
\end{equation}
We also have
\begin{equation}
    \label{A.1nabla2}
    \nabla_{D(s,a)}\nabla_{D(s',a')}\mathcal{L}(D,\pi_t) = 0, \quad \forall (s,a)\neq(s',a').
\end{equation}
From \eqref{A.1nabla1} and \eqref{A.1nabla2} we can get $\nabla_D\mathcal{L}(D,\pi_t)$ is a positive semi-definite matrix, i.e.,
$$\nabla^2_D\mathcal{L}(D,\pi_t) \succeq 0 .$$
That is to say, $\mathcal{L}(D, \pi_t)$ is a convex function w.r.t. $D$. Thus $\forall D_1,D_2\in(0,1)^{|\mathcal{S}|\times|\mathcal{A}|}$
\begin{align}
    \mathcal{L}(D_2, \pi_t) \geq \mathcal{L}(D_1, \pi_t) + \langle \nabla_{D_1} \mathcal{L}(D_1, \pi_t), D_2-D_1 \rangle .
    \label{2strongconvexProp}
\end{align}
Define that
\begin{equation}
D^\star = \arg\min_D\sum_{t=1}^T\mathcal{L}(D,\pi_t) ,
\end{equation}
and assign that $D_2=D^\star$ and $D_1=D_t$ in \eqref{2strongconvexProp} we can obtain
\begin{align}
\mathcal{L}(D^\star, \pi_t) \geq& \mathcal{L}(D_t, \pi_t) + \langle \nabla_{D_t} \mathcal{L}(D_t, \pi_t), D^\star-D_t \rangle \nonumber\\
=& \mathcal{L}(D_t, \pi_t) - \langle \nabla_{D_t} \mathcal{L}(D_t, \pi_t), D_t-D^\star \rangle \nonumber.
\end{align}
By transposition, we have
\begin{align}
- \langle \nabla_{D_t} \mathcal{L}(D_t, \pi_t), D_t-D^\star \rangle %\nonumber\\
&\leq \mathcal{L}(D^\star, \pi_t) - \mathcal{L}(D_t, \pi_t) \nonumber\\
&= -\left( \mathcal{L}(D_t, \pi_t) - \mathcal{L}(D^\star, \pi_t) \right) .
    \label{AppendixA2str-convex}
\end{align}
Then, we get
\begin{align}
    & \| D_{t+1} - D^\star \|_2^2\nonumber\\
    =& \| \Pi_{(0,1)^{|\mathcal{S}|\times|\mathcal{A}|}} \left( D_t - \omega_t\nabla_{D_t} \mathcal{L}(D_t, \pi_t) \right) - D^\star \|_2^2\nonumber \\
    \leq& \| D_t - \omega_t\nabla_{D_t} \mathcal{L}(D_t, \pi_t) - D^\star \|_2^2 \nonumber\\
    =& \| D_t - D^\star \|_2^2 - 2\omega_t\langle \nabla_{D_t} \mathcal{L}(D_t, \pi_t), D_t-D^\star \rangle \nonumber\\
    & + \omega_t^2\| \nabla_{D_t} \mathcal{L}(D_t, \pi_t) \|_2^2\nonumber\\
    \leq& \| D_t - D^\star \|_2^2 - 2\omega_t\left( \mathcal{L}(D_t, \pi_t)-\mathcal{L}(D^\star, \pi_t) \right) \nonumber\\
    & + \omega_t^2\| \nabla_{D_t} \mathcal{L}(D_t, \pi_t) \|_2^2 \quad (\text{from } \eqref{AppendixA2str-convex}) .
\label{AppendixASigFormula}
\end{align}
The first inequality holds for that: $(0,1)^{|\mathcal{S}|\times|\mathcal{A}|}$ is a convex set, $D_{t+1} = D_t - \omega_t\nabla_{D_t} \mathcal{L}(D_t, \pi_t)\in\mathbb{R}^{|\mathcal{S}|\times|\mathcal{A}|}$ and $D^\star \in (0,1)^{|\mathcal{S}|\times|\mathcal{A}|}$. By Pythagoras Theorem we have
\begin{align*}
&\|\left(D_t - \omega_t\nabla_{D_t} \mathcal{L}(D_t, \pi_t)\right) - D^\star\|_2 \\
\geq& \|\Pi_{(0,1)^{|\mathcal{S}|\times|\mathcal{A}|}} \left( D_t - \omega_t\nabla_{D_t} \mathcal{L}(D_t, \pi_t) \right) - D^\star\|_2 .
\end{align*}
Now, \eqref{AppendixASigFormula} can be simplified to
\begin{align}
    &\mathcal{L}(D_t, \pi_t) - \mathcal{L}(D^\star, \pi_t) \nonumber\\
    \leq& \frac{1}{2\omega_t}\left(\| D_t-D^\star \|_2^2 - \| D_{t+1}-D^\star \|_2^2\right) + \frac{\omega_t}{2}\| \nabla_{D_t} \mathcal{L}(D_t, \pi_t) \|_2^2 \nonumber\\
    \leq& \frac{1}{2\omega_t}\left(\| D_t-D^\star \|_2^2 - \| D_{t+1}-D^\star \|_2^2\right) + \frac{\omega_t}{2} G^2 .
    \label{AppendixA1}
\end{align}
After summing \eqref{AppendixA1} from $t=1$ to $T$ and define $\frac{1}{\omega_0}=0$, we can obtain
\begin{align}
\label{eq_regret_bound_process}
    &\sum_{t=1}^T \mathcal{L}(D_t, \pi_t) - \sum_{t=1}^T \mathcal{L}(D^\star, \pi_t) \nonumber\\
    \leq& \sum_{t=1}^T \frac{1}{2\omega_t}\left(\| D_t-D^\star \|_2^2 - \| D_{t+1}-D^\star \|_2^2\right) + \sum_{t=1}^T \frac{\omega_t}{2} G^2 \nonumber\\
    =& \sum_{t=1}^T \| D_t-D^\star \|_2^2\left( \frac{1}{2\omega_t}-\frac{1}{2\omega_{t-1}} \right) - \frac{1}{2\omega_T}\|D_{T+1}-D^\star\|_2^2 \nonumber\\
    & + \sum_{t=1}^T \frac{\omega_t}{2} G^2 \nonumber\\
    \leq& \sum_{t=1}^T \| D_t-D^\star \|_2^2\left( \frac{1}{2\omega_t}-\frac{1}{2\omega_{t-1}} \right) + \sum_{t=1}^T \frac{\omega_t}{2} G^2 \nonumber\\
    =& \| D_T-D^\star \|_2^2 \frac{1}{2\omega_T} + \sum_{t=1}^T \frac{\omega_t}{2} G^2 \nonumber\\
    =& \frac{\|D_T-D^\star\|_2^2 G \sqrt{T}}{2\Omega} + \frac{\Omega G}{2}\sum_{t=1}^T \frac{1}{\sqrt{t}} \quad (\omega_t = \frac{\Omega}{G\sqrt{t}}) \nonumber\\
    \leq &\frac{\|D_T-D^\star\|_2^2 G \sqrt{T}}{2\Omega} + \Omega G \sqrt{T} \quad (\sum_{t=1}^T\frac{1}{\sqrt{t}}\leq 2\sqrt{T}) \nonumber\\
    \leq& \frac{3\overline{\Omega} G \sqrt{T}}{2} ,
\end{align}
where $\overline{\Omega}=\max\{\frac{\|D_T-D^\star\|_2^2}{\Omega}, \Omega\}$. As a result, we can obtain the regret bound
\begin{align}
    \frac{1}{T} \sum_{t=1}^T \mathcal{L}(D_t, \pi_t) &\leq \frac{1}{T} \sum_{t=1}^T \mathcal{L}(D^\star, \pi_t) + \frac{3\overline{\Omega} G}{2\sqrt{T}} \nonumber\\
    &= \frac{1}{T} \min_D \sum_{t=1}^T \mathcal{L}(D, \pi_t) + \frac{3\overline{\Omega} G}{2\sqrt{T}}.
    \label{AppendixARegretBound}
\end{align}

For the left hand of \eqref{AppendixARegretBound}, we have
\begin{align}
    & \frac{1}{T} \sum_{t=1}^T \mathcal{L}(D_t, \pi_t) \nonumber\\
    \geq& \max_{\pi\in\{\pi_t| t\in[1,T]\}} \frac{1}{T} \sum_{t=1}^T \mathcal{L}(D_t, \pi) \quad (\pi_t=\arg\max_\pi \mathcal{L}(D_t,\pi)) \nonumber\\
    \geq& \max_{\pi\in\{\pi_t| t\in[1,T]\}} \mathcal{L}(\frac{1}{T} \sum_{t=1}^T D_t, \pi) \quad (\text{Jensen inequality}) \nonumber\\
    =& \max_{\pi\in\{\pi_t| t\in[1,T]\}} \mathcal{L}(\overline{D}, \pi).
    \label{AppendixAAdversial}
\end{align}
From \eqref{AppendixARegretBound} and \eqref{AppendixAAdversial} we can obtain
$$\max_{\pi\in\{\pi_t| t\in[1,T]\}}\mathcal{L}(\overline{D},\pi) \leq \frac{1}{T}\min_D\sum_{t=1}^T \mathcal{L}(D,\pi_{t}) + \frac{3\overline{\Omega} G}{2\sqrt{T}}.$$
And trivially, 
\[
\frac{1}{T}\sum_{t=1}^T\mathcal{L}(\overline{D}, \pi_t) \leq \max_{\pi\in\{\pi_t| t\in[1,T]\}}\mathcal{L}(\overline{D},\pi).
\]
Therefore, we finally get
\[
\frac{1}{T}\sum_{t=1}^T\mathcal{L}(\overline{D}, \pi_t) \leq \frac{1}{T}\min_D\sum_{t=1}^T \mathcal{L}(D,\pi_{t}) + \frac{3\overline{\Omega} G}{2\sqrt{T}}.
\]
\end{proof}

\subsection{Proof of Theorem~\ref{thm_vclub}}
\label{supp_theorem_proof}
\begin{theorem}[Variational Contrastive Log-Ratio Upper Bound~\cite{cheng2020vclub}]
Let $q(s'| z;\theta)$ be a variation approximation of $p(s'| z)$ with parameter $\theta$.  Denote $q(z,s';\theta)=q(s'| z;\theta)p(z)$. If 
\begin{equation}
\label{eq_club_constraint_supp}    
D_{KL}(p(z,s') \Vert q(z,s';\theta))\leq D_{KL}(p(s')p(z)\Vert q(z,s';\theta)),
\end{equation}
then $I(z;s') \leq I_\text{vCLUB}(z;s')$, where
\begin{equation}
\label{eq_club_upper_bound_supp}    
I_\text{vCLUB}= \mathbb{E}_{p(z,s')}\left[\log q(s'| z;\theta)\right]-\mathbb{E}_{p(z)}\mathbb{E}_{p(s')}\left[\log q(s'| z;\theta)\right].
\end{equation}

\end{theorem}
\begin{proof}
We have
\[
\begin{aligned}
D_{KL}(p(z,s') \Vert q(z,s';\theta))&=\mathbb{E}_{p(z,s')}\left[\log{\frac{p(z,s')}{q(z,s';\theta)}}\right],\\
D_{KL}(p(s')p(z)\Vert q(z,s';\theta)) &= \mathbb{E}_{p(z)}\mathbb{E}_{p(s')}\left[\log{\frac{p(s')p(z)}{q(z,s';\theta)}}\right].
\end{aligned}
\]
Thus, we can convert \eqref{eq_club_constraint_supp} to
\begin{equation}
\mathbb{E}_{p(z,s')}\left[\log{\frac{p(z,s')}{q(z,s';\theta)}}\right]\leq\mathbb{E}_{p(z)}\mathbb{E}_{p(s')}\left[\log{\frac{p(s')p(z)}{q(z,s';\theta)}}\right].
\label{AppendixA2}
\end{equation}
By substituting $q(z,s';\theta)$ with $q(s'| z;\theta)p(z)$, we have
\[
\begin{aligned}
\mathbb{E}_{p(z,s')}\left[\log{\frac{p(z,s')}{q(z,s';\theta)}}\right]&\leq\mathbb{E}_{p(z)}\mathbb{E}_{p(s')}\left[\log{\frac{p(s')p(z)}{q(z,s';\theta)}}\right]\\
\mathbb{E}_{p(z,s')}\left[\log{\frac{p(s'| z)p(z)}{q(s'| z;\theta)p(z)}}\right]&\leq\mathbb{E}_{p(z)}\mathbb{E}_{p(s')}\left[\log{\frac{p(s')p(z)}{q(s'| z;\theta)p(z)}}\right]\\
\mathbb{E}_{p(z,s')}\left[\log{\frac{p(s'| z)}{q(s'| z;\theta)}}\right]&\leq\mathbb{E}_{p(z)}\mathbb{E}_{p(s')}\left[\log{\frac{p(s')}{q(s'| z;\theta)}}\right],
\end{aligned}
\]
which is equivalent to
\begin{equation}
\label{eq_supp_final_inequality}
\begin{aligned}
&~~~~\mathbb{E}_{p(z,s')}\log{p(s'| z)} - \mathbb{E}_{p(z,s')}\log{p(s'| z;\theta)}\\
&\leq \mathbb{E}_{p(s')}[\log{p(s')}] -  \mathbb{E}_{p(z)}\mathbb{E}_{p(s')}\log{q(s'| z;\theta)}\\
\end{aligned}
\end{equation}
Thus, we have
\[
\begin{aligned}
&~~~~\mathbb{E}_{p(z,s')}\log{p(s'| z)} - \mathbb{E}_{p(s')}\log{p(s')} \\
&\leq  \mathbb{E}_{p(z,s')}\log{p(s'| z;\theta)} -  \mathbb{E}_{p(z)}\mathbb{E}_{p(s')}\log{q(s'| z;\theta)}.\\
\end{aligned}
\]
Recall that the mutual information is 
\[
I(z;s') = \mathbb{E}_{p(z,s')}[\log p(s'| z)] - \mathbb{E}_{p(s')}[\log(p(s'))]. 
\]
\eqref{eq_supp_final_inequality} can induce that 
\[
\begin{aligned}
I(z;s') &= \mathbb{E}_{p(z,s')}[\log{p(s'| z)}] - \mathbb{E}_{p(s')}[\log{p(s')}]\\
&\leq \mathbb{E}_{p(z,s')}\log{p(s'| z;\theta)} - \mathbb{E}_{p(z)}\mathbb{E}_{p(s')}\log{q(s'| z;\theta)}\\
&= I_\text{vCLUB}(z;s').
\end{aligned}
\]
\end{proof}
\section{Experiment Details}
\label{supp_experiment_details}
In this part, we introduce the experiment details, including \textit{expert data} (\ref{sec_expert_data_description}), \textit{transfer task constructions} (\ref{sec_transfer_task_construct}), \textit{baseline implementations} (\ref{sec_baseline_implementation}), and \textit{hyper-parameters settings} (\ref{sec_hyper_parameter_setting}).

\subsection{Expert Data}
\label{sec_expert_data_description}

We use PPO~\cite{schulman2017ppo} to learn a policy from scratch for each environment. The deterministic learned policy is then utilized to sample several trajectories, which serve as expert demonstrations. Detailed information regarding these expert demonstrations is listed in \cref{table_supp_expert} and \ref{table_supp_expert_rampti}.

\subsection{Transfer Task Constructions}
\label{sec_transfer_task_construct}
\begin{table}[ht]
    \centering
    \caption{Sampling range coefficients of the environment parameters.}
    \label{table_supp_envrionment_range}
    \begin{tabular}{lr}\toprule
\textbf{Environment} & $\xi$ \\\midrule
Ant-v2 & \multirow{5}{*}{$1.5$} \\
HalfCheetah-v2& \\
Hopper-v2& \\
Walker2d-v2& \\
Humanoid-v2 & \\\midrule
EnergyLimAnt-v2&\multirow{4}{*}{$0.5$}\\
EnergyLimHalfCheetah-v2\\
EnergyLimHopper-v2 \\
EnergyLimWalker2d-v2\\
\bottomrule
% \bottomrule
\end{tabular}
\end{table}

\begin{table*}[t]
    \centering
    \caption{Expert demonstration information for the tasks with the default reward functions. The expert policy is deterministic.}
    \label{table_supp_expert}
    \resizebox{\textwidth}{!}{
    \begin{tabular}{l|cccccc}\toprule
&Ant-v2&HalfCheetah-v2&Hopper-v2&Walker2d-v2&Humanoid-v2&DisableAnt-v2\\\midrule
Average return&$2958.42$ &$3930.16$&$3315.92$&$3272.10$&$6623.60$&$1081.91$\\
Number of state-action pairs &$8,000$&$8,000$&$8,000$&$8,000$&$80,000$&$4,000$\\
Number of trajectories &$8$&$8$&$8$&$8$&$80$&$8$\\
\bottomrule
\end{tabular}
}
\end{table*}

\begin{table*}[t]
    \centering
    \caption{Expert demonstration information for the tasks with the energy-limit reward functions. The expert policy is deterministic.}
    \label{table_supp_expert_rampti}
    \resizebox{\textwidth}{!}{
\begin{tabular}{l|cccc}\toprule
&EnergyLimAnt-v2&EnergyLimHalfCheetah-v2&EnergyLimHopper-v2&EnergyLimWalker2d-v2\\\midrule
Average return&$3734.42$ &$1917.46$&$2434.99$&$3080.34$\\
Number of state-action pairs &$25,000$&$25,000$&$25,000$&$25,000$\\
Number of trajectories &$25$&$25$&$25$&$25$\\
\bottomrule
\end{tabular}
}
\end{table*}
We construct the dynamics changes based on an open-sourced repository\footnote{https://github.com/dennisl88/rand\_param\_envs}, similar to the implementation of PEARL~\cite{rakelly2019pearl}\footnote{https://github.com/katerakelly/oyster}~\citep{rakelly2019pearl}. Given a sampling range factor for the dynamics parameters, $\xi$, we sample a new dynamics parameter as follows:
\[
\text{new\_param} = 1.5^{\texttt{uniform}(-\xi, \xi)} \text{param}_\text{origin},
\]
where $\texttt{uniform}(-\xi, \xi)$ denotes sampling a random variable uniformly from the interval $[-\xi,\xi]$, and $\text{param}\text{origin}$ is the original parameter. For example, in \texttt{HalfCheetah} with \texttt{gravity}-transfer, f $\xi=1.5$ and $\text{param}\text{origin}=-9.81$, the new gravity parameter is sampled as:
\[
\text{new\_gravity} = 1.5^{\texttt{uniform}(-1.5,1.5)}\times -9.81.
\]
The setting of $\xi$ for each environment is listed in \cref{table_supp_envrionment_range}. We use a smaller $\xi$ for the energy-limit tasks because even an \textit{oracle} cannot learn a high-return policy when $\xi$ is set to $1.5$.

For the dynamics transfer in the \texttt{DisableAnt} environment, we adopt the implementations from previous methods~\cite{fu2017airl,ni2020firl}\footnote{\url{https://github.com/twni2016/f-IRL/blob/main/envs/ant_env.py}}.
\subsection{Baseline Implementations}

\label{sec_baseline_implementation}
\begin{itemize}
    \item \textit{GAIL}~\cite{ho2016gail}. We implement GAIL as described in the original paper\cite{ho2016gail} using PPO as the RL optimization method. A lot of implementation details are taken from the \textit{Imitation Learning Baseline Implementations}~\cite{gleave2022imitation}\footnote{https://github.com/HumanCompatibleAI/imitation}.
    \item \textit{GAIL-B}. We introduce a replay buffer to GAIL to mitigate the policy dependency issue, referring to this variant as \textit{GAIL-B} (GAIL-Buffer). The buffer stores all generated data during training, and in each iteration, GAIL-B samples a batch of data from the replay buffer as generated data. For fair comparisons, the batch size is set to the number of samples collected by the policy in each iteration. Note that replay buffers have been used in DAC\cite{kostrikov2019dac} and ValueDice~\cite{kostrikov2020valuedice} to improve data efficiency in AIL.
    \item \textit{GAIL-DAC}. We extend GAIL-B by incorporating the absorbing state concept from DAC~\cite{kostrikov2019dac} and modifying the terminal signal accordingly\footnote{https://github.com/google-research/google-research/blob/master/dac}. To eliminate reward bias, at terminal states, GAIL-DAC transitions the current state to an absorbing state that always transitions to itself.
    \item \textit{AIRL-SA}~\cite{fu2017airl}. AIRL-SA aims to learn a dynamics-agnostic reward function by altering the discriminator architecture and reward function formulation. We introduce a potential function $h(s)$ and represent the discriminator network as $f_{\text{AIRL}}(s,a,s')=g(s,a)+\gamma h(s')-h(s)$, where $g(s,a)$ and $h(s)$ are neural networks that receive state-action and state as inputs. The discriminator is formalized as 
    \begin{equation}
    \label{eq_airl_discriminator}
        D_\text{AIRL}(s,a,s') = \frac{\exp\left\{f_\text{AIRL}(s,a,s')\right\}}{\exp\left\{f_\text{AIRL}(s,a,s')\right\}+\pi(a| s)}.
    \end{equation}
    The loss of the discriminator is also in the form of \eqref{eq_gail_loss}. 
    \item \textit{AIRL-SO}~\cite{fu2017airl}. Unlike AIRL-SA, AIRL-SO does not learn an action-dependent discriminator: $f_{\text{AIRL-SO}}(s,a,s')=g(s)+\gamma h(s')-h(s)$, where both $g(s)$ and $h(s)$ receive state as input. The formation of $g(s)$ is the only difference between AIRL-SO and AIRL-SA.
    \item \textit{VAIL}~\cite{peng2019vdb}. VAIL uses the principle of the \textit{information bottleneck} to constrain the mutual information between the input and the hidden layer output of the discriminator. Practically, we introduce two networks, $E_\text{VAIL}^\mu(s,a):\mathcal{S}\times\mathcal{A}\rightarrow \mathcal{Z}$ and $E_\text{VAIL}^\sigma(s,a):\mathcal{S}\times\mathcal{A}\rightarrow \mathcal{Z}$, to encode the state-action pair to a Gaussian distribution $\mathcal{N}(z| E_\text{VAIL}^\mu(s,a),\left(E_\text{VAIL}^\sigma(s,a)\right)^2)$. From this distribution, we sample a latent variable $z$ and input $z$ to a discriminator. The encoders are trained to minimize the discriminator loss and the KL divergence of the Gaussian distribution from a prior distribution $p(z)$, which is a standard Gaussian distribution.
    \item \textit{$f$-IRL}~\cite{ni2020firl}. This state-only IRL method aims to recover a stationary reward by state marginal matching under $f$-divergence. We use the official implementation of $f$-IRL\footnote{https://github.com/twni2016/f-IRL}.
    \item \textit{MCE-IRL}~\cite{ziebart2008maxent}. Maximum causal entropy IRL, a typical maximum entropy IRL method instantiated by a neural network. Our implementation of MCE-IRL is based on the codebase of the $f$-IRL repository. We modify the input of the discriminator from state-only to state-action.
    \item \textit{IQ-Learn}~\cite{garg2021iql}. Inverse soft-Q learning, a non-adversarial imitation learning method applicable to IRL scenarios. We use the official implementation of IQ-Learn\footnote{https://github.com/Div99/IQ-Learn}.
    \item \textit{transferred policy}. This baseline directly transfers the learned policy of GAIL to a new environment without parameter adjustments.
    \item \textit{oracle}. This baseline trains policies using true environment reward functions via PPO~\cite{schulman2017ppo}.
\end{itemize}

\begin{table*}[htbp]
    \centering
    \caption{Return $\pm$ standard error of the learned policy via AIL in original environments. Instead of training a policy with the learned reward from scratch, we present the performance of the policies learned using AIL. All baselines are repeated with six seeds.}
    \label{tab_reward_origin_env}
    \resizebox{\textwidth}{!}{
    \begin{tabular}{l|r@{~$\pm$~}lr@{~$\pm$~}lr@{~$\pm$~}lr@{~$\pm$~}lr@{~$\pm$~}lr@{~$\pm$~}lr@{~$\pm$~}lc}\toprule
& \multicolumn{2}{c}{DARL} & \multicolumn{2}{c}{GAIL-DAC} & \multicolumn{2}{c}{VAIL} & \multicolumn{2}{c}{GAIL-B} & \multicolumn{2}{c}{GAIL} & \multicolumn{2}{c}{AIRL-SO} & \multicolumn{2}{c}{AIRL-SA} & expert\\\midrule
Ant-v2 & $ {3202} $ & $ {16} $ & $2462$& $230$ & $3132$& $133$ & $2379$& $200$ & $3030$& $44$ & $2809$& $20$ & $3046$& $54$ & $2958$\\
HalfCheetah-v2 & $4163$& $203$ & $3656$& $542$ & $ {4745} $ & $ {188} $ & $4216$& $341$ & $4496$& $116$ & $3472$& $300$ & $4048$& $81$ & $3930$\\
Hopper-v2 & $3057$& $65$ & $2882$& $152$ & $3109$& $21$ & $2767$& $324$ & $3052$& $210$ & $2186$& $311$ & $2602$& $426$ & $ {3316} $ \\
Walker2d-v2 & $2566$& $327$ & $3151$& $445$ & $3948$& $98$ & $ {4133} $ & $ {97} $ & $3365$& $446$ & $3010$& $325$ & $3200$& $54$ & $3272$\\
Humanoid-v2 & $5620$& $906$ & $5907$& $194$ & $5770$& $426$ & $6009$& $220$ & $6203$& $121$ & $381$& $34$ & $ {374} $ & $ {45} $ & $6624$\\
DisableAnt-v2 & $950$& $20$ & $741$& $33$ & $867$& $44$ & $748$& $37$ & $874$& $11$ & $ {2} $ & $ {3} $ & $22$& $13$ & $1082$\\
EnergyLimAnt-v2 & $ {3918} $ & $ {83} $ & $3541$& $52$ & $3727$& $79$ & $3386$& $193$ & $3802$& $89$ & $-666$& $176$ & $3734$& $81$ & $3734$\\
EnergyLimHalfCheetah-v2 & $947$& $74$ & $837$& $303$ & $1586$& $51$ & $434$& $303$ & $1413$& $56$ & $-496$& $304$ & $1501$& $49$ & $ {1917} $ \\
EnergyLimHopper-v2 & $2128$& $266$ & $2113$& $276$ & $2180$& $53$ & $2324$& $28$ & $2304$& $27$ & $1180$& $211$ & $1532$& $302$ & $ {2435} $ \\
EnergyLimWalker2d-v2 & $2056$& $347$ & $2318$& $163$ & $2375$& $352$ & $2533$& $242$ & $2502$& $58$ & $-942$& $119$ & $399$& $345$ & $ {3080} $ \\
\bottomrule
    \end{tabular}
    }
\end{table*}

\begin{table*}[htbp]
    \centering
    \caption{Return $\pm$ standard error of the learned policy via IRL baselines in original environments. Instead of training a policy with the learned reward from scratch, we present the performance of the policies learned by IRL. All baselines are repeated with six seeds.}
    \label{tab_reward_origin_env_other_methods}
    \begin{tabular}{l|r@{~$\pm$~}lr@{~$\pm$~}lr@{~$\pm$~}lr}\toprule
& \multicolumn{2}{c}{$f$-IRL} & \multicolumn{2}{c}{MCE-IRL} & \multicolumn{2}{c}{IQ-Learn} & expert\\\midrule
Ant-v2 & $3167$& $37$ & $ {3422} $ & $ {22} $ & $770$& $100$ & $2958$\\
HalfCheetah-v2 & $4725$& $92$ & $4581$& $33$ & $ {7187} $ & $ {898} $ & $3930$\\
Hopper-v2 & $3286$& $9$ & $ {3337} $ & $ {5} $ & $1652$& $164$ & $3316$\\
Walker2d-v2 & $ {3797} $ & $ {128} $ & $3737$& $119$ & $3722$& $243$ & $3272$\\
Humanoid-v2 & $6442$& $153$ & $2958$& $1235$ & $ {1207} $ & $ {263} $ & $6624$\\
DisableAnt-v2 & $ {1070} $ & $ {168} $ & $1151$& $5$ & $739$& $168$ & $1082$\\
EnergyLimAnt-v2 & $-2890$& $1445$ & $3229$& $174$ & $1194$& $517$ & $ {3734} $ \\
EnergyLimHalfCheetah-v2 & $1511$& $215$ & $1436$& $83$ & $1867$& $46$ & $ {1917} $ \\
EnergyLimHopper-v2 & $2203$& $38$ & $2057$& $346$ & $253$& $871$ & $ {2435} $\\
EnergyLimWalker2d-v2 & $605$& $122$ & $1334$& $253$ & $-1493$& $1426$ & $ {3080} $ \\
\bottomrule
    \end{tabular}
\end{table*}
To evaluate the performance of the implemented baselines, we first test the methods in their original environments under the IRL setting. The environment returns of the learned policies by AIL are listed in \cref{tab_reward_origin_env} and \cref{tab_reward_origin_env_other_methods}. Consistent with prior work~\cite{ni2020firl,zeng2022mlirl}, we observe that AIRL exhibits instability during training and struggles to achieve high returns in complex environments like \texttt{Humanoid}. Although AIRL-SO fails to obtain high-return policies in \texttt{DisableAnt}, its reward function achieves the highest return among the GAIL-variant baselines in the transfer task. This suggests that a method's imitation ability does not directly reflect the quality of the learned reward. Furthermore, we find the training process of IQ-Learn to be highly unstable, as noted in~\cite{zeng2022mlirl}. For instance, IQ-Learn reaches a return of $7,187$ in \texttt{HalfCheetah} but only $770$ in \texttt{Ant}, representing the highest and lowest returns among all baselines, respectively. The training curves of IQ-Learn also exhibit significant jittering.

Another notable observation is that introducing a memory buffer can impact imitation optimality. This is evident when comparing the returns of GAIL with its variants that use memory buffers (GAIL-B and GAIL-DAC). In $7/10$ tasks, GAIL-B, and in $10/10$ tasks, GAIL-DAC perform worse than GAIL, possibly due to the buffers affecting GAIL's optimality, as also highlighted in~\cite{kostrikov2020valuedice}. Additionally, AIRL-SO underperforms compared to AIRL-SA in all energy-limited environments. This is attributed to AIRL-SO omitting action costs during imitation. 

An interesting result is that $f$-IRL achieves relatively high returns in \texttt{EnergyLimHalfCheetah} and \texttt{EnergyLimHopper}, despite learning a state-only reward. This implies that in these environments, the state and action are coupled such that recovering the expert state-action distribution can be approximated by recovering the expert state distribution. Conversely, \texttt{EnergyLimAnt} and \texttt{EnergyLimWalker2d} do not exhibit this feature, resulting in $f$-IRL's failure to achieve high returns in these environments, with a particularly low return of $-2,890$ in \texttt{EnergyLimAnt}.
\subsection{Hyper-parameters}
\label{sec_hyper_parameter_setting}
We list the hyper-parameters shared across various baselines in \cref{table_supp_shared_parameters}. The specific hyper-parameters for \ours, GAIL, AIRL, and VAIL are detailed in \cref{table_supp_ours_parameters}, \ref{table_supp_gail_parameters}, \ref{table_supp_airl_parameters}, and \ref{table_supp_vail_parameters}, respectively. Additionally, g\_steps and d\_steps are tuned for different tasks, as listed in \cref{table_supp_shared_parameters_g_steps}. In MuJoCo tasks, g\_steps are typically set to $1$ as established by~\cite{gleave2022imitation}. However, for \texttt{Humanoid} and energy-limited tasks, which are more challenging for all GAIL variants, we increase g\_steps accordingly. For environments other than \texttt{Humanoid}, we set the g\_steps of \ours to be $5$ times larger than other baselines to approximate the $\arg\max$ in line~\ref{alg_ogd_update_g} of Alg.~\ref{alg_OGD}. We also increase the d\_steps of variants that maintain a memory buffer for the discriminator to $2$. This adjustment is necessary because GAIL-B and GAIL-DAC require sampling a batch of data from the memory buffer first when training the discriminator. The sampled data might correspond to policies far from the current policy. Without additional training steps, the discriminator may delay recognizing the current policy's behavior, thus affecting training efficiency. The hyper-parameters of MCE-IRL, $f$-IRL, and IQ-Learn are set to their default values as specified in their respective codebases for each environment. % We will introduce the settings of $\eta$ in \cref{sec_eta_searching}.
\begin{table}[ht]
    \centering
    \caption{Shared hyper-parameters for GAIL variants and \ours.}
    \label{table_supp_shared_parameters}
    \begin{tabular}{lr}\toprule
\textbf{Attribute} & \textbf{Value}\\\midrule
Policy lr& $3e-4$\\ 
Value lr& $1.5e-4$\\ 
Discriminator lr & $5e-4$ \\ 
Discount factor $\gamma$ & $0.99$ \\ 
PPO clip factor & $0.2$ \\ 
GAE $\lambda$ & $0.95$\\ 
Number of PPO epochs & $10$ \\ 
Number of PPO mini-batches & $5$ \\ 
Number of samples per iteration & $10,000$ \\ 
Policy and value optimizers & Adam \\ 
Discriminator optimizer & RMSProp \\ 
Hidden layers of the policy network  &  $[256, 128]$\\ 
Activations of the policy network &  [ReLU, ReLU, Tanh]\\ 
Hidden layers of the value network &  $[256, 128]$\\ 
Activations of the value network &  [ReLU, ReLU, Linear]\\ 
Discriminator buffer size & $5e6$ (GAIL-B and GAIL-DAC)\\
\bottomrule
\end{tabular}
\end{table}

\begin{table}[htbp]
    \centering
    \caption{g\_steps and d\_steps for GAIL variants and \ours.}
    \label{table_supp_shared_parameters_g_steps}
    \begin{tabular}{llr}\toprule
\textbf{Attribute} & \textbf{Method} & \textbf{Value}\\\midrule
\multirow{5}{*}{g\_steps} & All methods in \texttt{Humanoid}  & $10$  \\ 
 & \ours in energy-limited envs.  & $10$  \\ 
 & \ours in other envs. & $5$   \\
 & Other GAIL variants in energy-limited envs.  & $2$  \\ 
 & Otherwise & $1$  \\
\midrule
\multirow{2}{*}{d\_steps} &  GAIL-B, GAIL-DAC & $2$  \\
                          &  Otherwise   & $1$\\
\bottomrule
\end{tabular}
\end{table}

\begin{table}[htbp]
    \centering
    \caption{Particular hyper-parameters for \ours.}
    \label{table_supp_ours_parameters}
    \begin{tabular}{lr}\toprule
\textbf{Attribute} & \textbf{Value}\\\midrule
Hidden layers of the state encoder $E_s(s)$ &  $[128, 128]$\\ 
Activations of the state encoder $E_s(s)$  &  [ReLU, ReLU, Tanh]\\ 
Hidden layers of the action encoder $E_a(a)$ &  $[128, 128]$\\ 
Activations of the action encoder $E_a(a)$ &  [ReLU, ReLU, Tanh]\\ 
Hidden layers of the discriminator $D(z)$ &  $[128, 128]$\\ 
Activations of the discriminator $D(z)$ &  [ReLU, ReLU, Sigmoid]\\ 
State embedding dimensions $\#\text{Dim}(\mathcal{Z}_s)$ & $32$ \\ 
Action embedding dimensions $\#\text{Dim}(\mathcal{Z}_a)$ & $8$ \\ 
Discriminator backing up interval $H$ & $1$ ($10$ for \texttt{Humanoid})  \\ 
Transition training mini-batch size & $512$ \\ 
Transition learning rate & $1e-3$ \\  %\midrule
Transition update steps & 50 \\
Maximum output of the discriminator $c$ & 0.5\\
\bottomrule
\end{tabular}
\end{table}

\begin{table}[htbp]
    \centering
    \caption{Particular hyper-parameters for GAIL.}
    \label{table_supp_gail_parameters}
    \begin{tabular}{lr}\toprule
\textbf{Attribute} & \textbf{Value}\\\midrule
Hidden layers of the discriminator  &  $[128, 128, 128]$\\ 
Activations of the discriminator &  [ReLU, ReLU, ReLU, Sigmoid]\\
\bottomrule
\end{tabular}
\end{table}

\begin{table}[htbp]
    \centering
    \caption{Particular hyper-parameters for AIRL-SA and AIRL-SO.}
    \label{table_supp_airl_parameters}
    \begin{tabular}{lr}\toprule
\textbf{Attribute} & \textbf{Value}\\\midrule
Hidden layers of the reward function $g$&  $[128, 256, 128]$\\ 
Activations of the reward function  $g$&  [ReLU, ReLU, ReLU, Linear]\\ 
Hidden layers of the potential function $h$ &  $[128, 256, 128]$\\ 
Activations of the potential function $h$ &  [ReLU, ReLU, ReLU, Linear]\\
\bottomrule
\end{tabular}
\end{table}

\begin{table}[htbp]
    \centering
    \caption{Particular hyper-parameters for VAIL.}
    \label{table_supp_vail_parameters}
    \begin{tabular}{lr}\toprule
\textbf{Attribute} & \textbf{Value}\\\midrule
Hidden layers of $E_\text{VAIL}^\mu(s,a)$  &  $[256, 128]$\\ 
Activations of $E_\text{VAIL}^\mu(s,a)$  &  [ReLU, ReLU, Linear]\\ 
Hidden layers of $E_\text{VAIL}^\sigma(s,a)$ &  $[256, 128]$\\ 
Activations of $E_\text{VAIL}^\sigma(s,a)$ &  [ReLU, ReLU, Linear]\\ 
Hidden layers  of the discriminator $D(z)$&  $[]$\\ 
Activations of the discriminator $D(z)$ &  [Sigmoid]\\ 
KL factor $\beta$ & adaptive $\beta (I_c=0.5)$\\ 
Embedding dimension & 128\\
\bottomrule
\end{tabular}
\end{table}

\begin{table}[htbp]
    \centering
    \caption{Values of $\eta$ and the corresponding average transition loss in \ours. `$\eta\times$ tran. loss' represents the product of the transition loss and $\eta$. The subscript $\times 4$ indicates that in \texttt{Ant}, $75\%$ of the state dimensions are always zero. Therefore, the transition loss for the non-zero dimensions is multiplied by $4$. Despite $\eta$ varying from $0.02$ to $0.25$, the transition loss ranges from $0.282$ to $16.415$, while the $\eta$-transition-loss products generally fall within a narrow range of $[0.1, 0.44]$. }
    \label{table_supp_ours_eta_setting}
    \begin{tabular}{lr@{~$\pm$~}lcc}\toprule
\textbf{Environment} & \multicolumn{2}{c}{\textbf{Transition loss}} & \textbf{$\eta$} & $\eta\times$ tran. loss\\\midrule
Ant-v2 &$0.357$& $0.022$ & $0.1$ & $0.04_{\times 4=0.16}$\\
EnergyLimAnt-v2&$0.282$& $0.015$ & $0.1$ & $0.03_{\times 4=0.12}$\\
DisableAnt-v2 &$ {0.389} $ & $ {0.035} $& $0.25$ & $0.10_{\times 4=0.40}$\\\midrule
HalfCheetah-v2&$16.415$& $0.840$ & $0.02$ & $0.33$\\
Hopper-v2&$1.678$& $0.458$ & $0.2$& $0.34$\\
Walker2d-v2&$4.410$& $0.402$  & $0.1$ & $0.44$\\
EnergyLimHalfCheetah-v2 &$5.105$& $0.362$ & $0.02$ & $0.10$ \\
EnergyLimHopper-v2&$1.318$& $0.322$ & $0.08$& $0.11$\\
EnergyLimWalker2d-v2 &$4.399$& $0.390$& $0.08$& $0.35$\\
\bottomrule
\end{tabular}
\end{table}

The values of $\eta$ in \ours\ significantly influence policy performance, particularly in transfer scenarios, as demonstrated in the ablation studies (\cref{fig_ablation_on_eta}). To determine the optimal $\eta$ for each task, we conducted a grid search. From \cref{fig_ablation_on_eta}, $\eta = 0.1$ emerged as the optimal value for $\texttt{Ant}$, leading us to set the search range to $\{0.02, 0.08, 0.1, 0.2, 0.25\}$. The chosen $\eta$ values for each task are detailed in \cref{table_supp_ours_eta_setting}.
While $\eta$ varies across different environments, we observed that the optimal $\eta$ is closely related to the transition loss of the transition network, defined as $\frac{\Vert s' - \tilde{s}' \Vert_2^2}{ \text{Dim}(s)}$, where $s'$ is the actual next state, $\tilde{s}'$ is the predicted next state by the transition network $q$, and $\text{Dim}(s)$ is the dimension of $s$. We also present the products of $\eta$ and corresponding transition losses in \cref{table_supp_ours_eta_setting}.
In the \texttt{Ant} environment and its variants, approximately 75\% of state dimensions are zero. Consequently, when calculating the products, we multiplied the transition losses in these environments by 4 to account for the non-zero dimensions. In \cref{table_supp_ours_eta_setting}, $\eta$ ranges from 0.02 to 0.25, and transition losses range from 0.282 to 16.415. Although the upper bounds of $\eta$ and transition loss are significantly larger than their lower bounds by factors of 11 and 57, respectively, the $\eta$-transition-loss products fall within the range of [0.1, 0.44]. The upper bound of these products is only 3.4 times larger than the lower bound, indicating no magnitude difference.

This observation implies that environments with high transition losses typically require a smaller $\eta$, and vice versa. For instance, in the \texttt{Humanoid} environment, where the transition function is the most challenging to approximate, we chose a very small $\eta$ of $5 \times 10^{-4}$ as the regularization factor.

\section{More Experiment Results}
\label{supp_more_exp}

\begin{figure}[ht]
    \centering
    \includegraphics[width=0.6\linewidth]{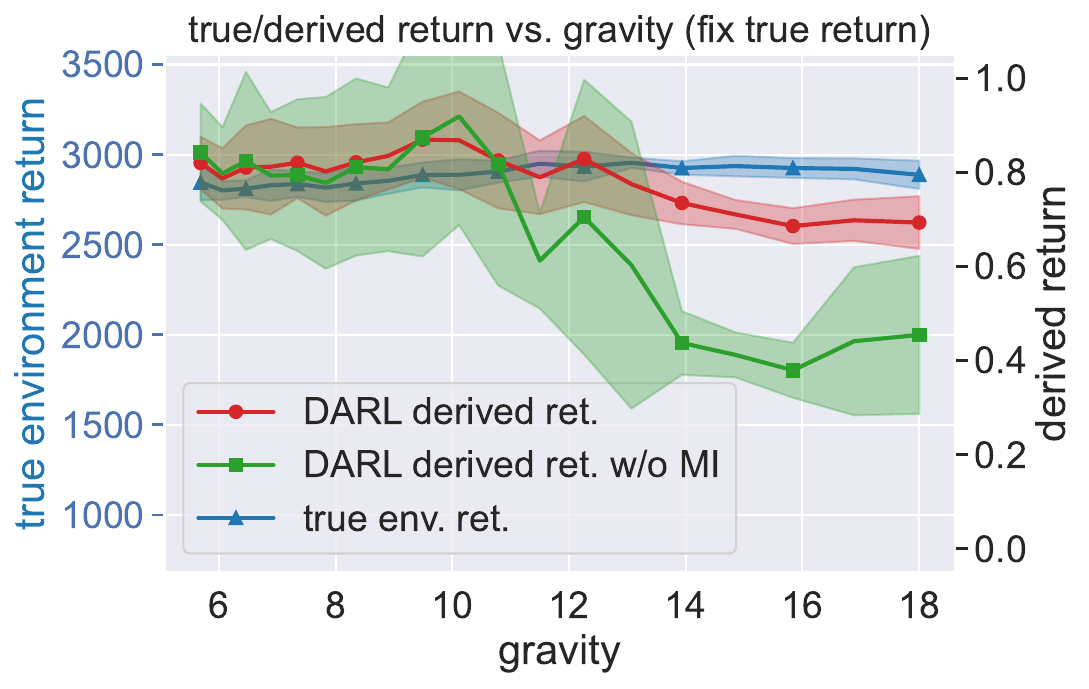}
    \caption{True environment returns and derived returns vs. \texttt{gravity}.  We trained a policy in \texttt{HalfCheetah} under varying \texttt{gravity} perturbations, collecting data throughout the training process. Trajectories achieving at least $75\%$ of the expert return were selected to illustrate their derived returns.}
    \label{fig_dynamics_gravity_curve_fix_ret_long}
\end{figure}
\begin{figure*}[ht]
    \centering
    \includegraphics[width=0.7\linewidth]{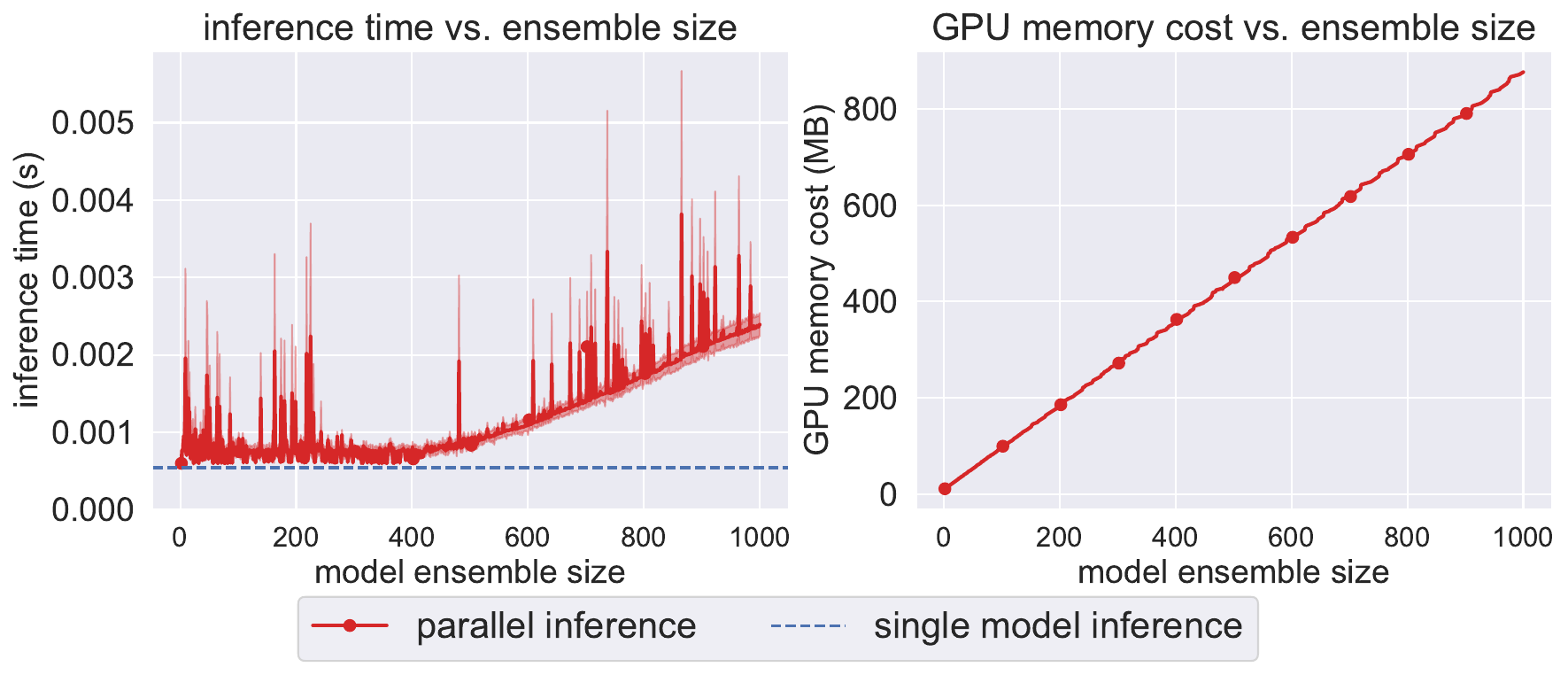}
    \caption{Time and GPU memory cost vs. model ensemble size. Inference time and memory cost are measured while predicting rewards for $128$ state-action pairs in \texttt{Ant}. We accelerate the computation by GPU parallelization. \textit{Single model inference} indicates the average inference time of a single discriminator.}
    \label{fig_time_cost}
\end{figure*}
\begin{figure}[ht]
    \centering
    \includegraphics[width=0.32\linewidth]{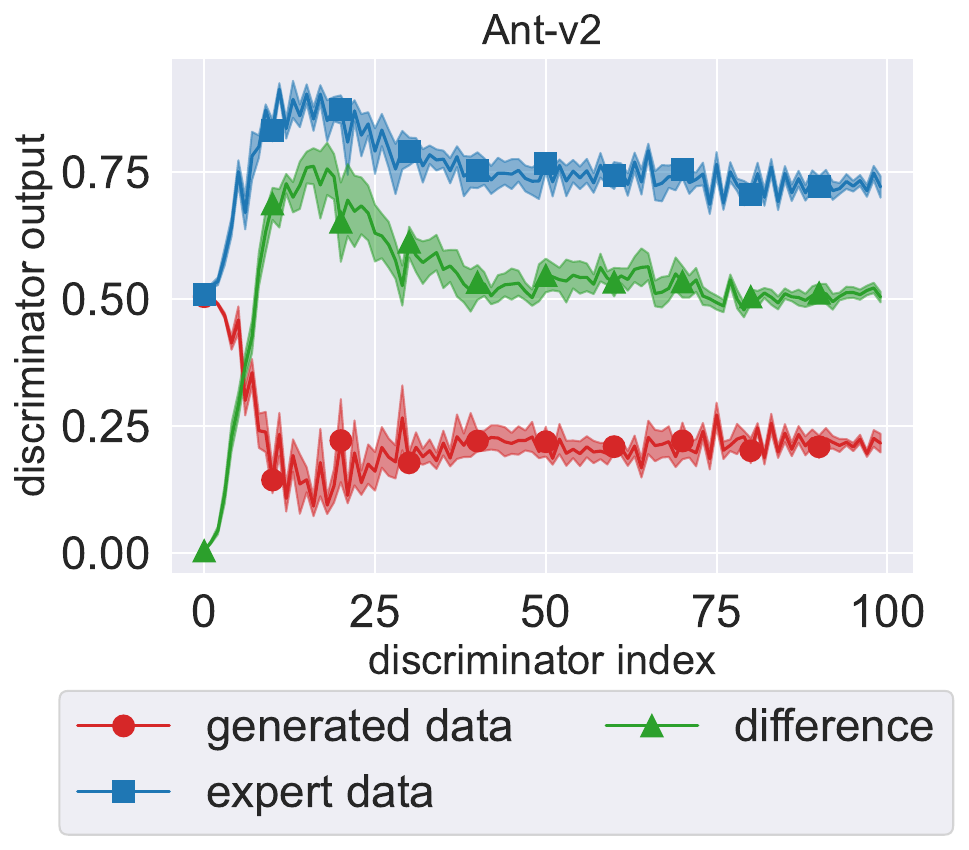}
    \caption{Average outputs from historical discriminators applied to both generated and expert data. For the $i$-th discriminator, the generated data is obtained by sampling according to the policy acquired at the $i$-th iteration. The \textit{difference} curve represents the difference between the discriminator outputs for expert data and generated data.}
    \label{fig_changing_chara}
\end{figure}
 
\begin{figure}[ht]
    \centering
    \includegraphics[width=0.5\linewidth]{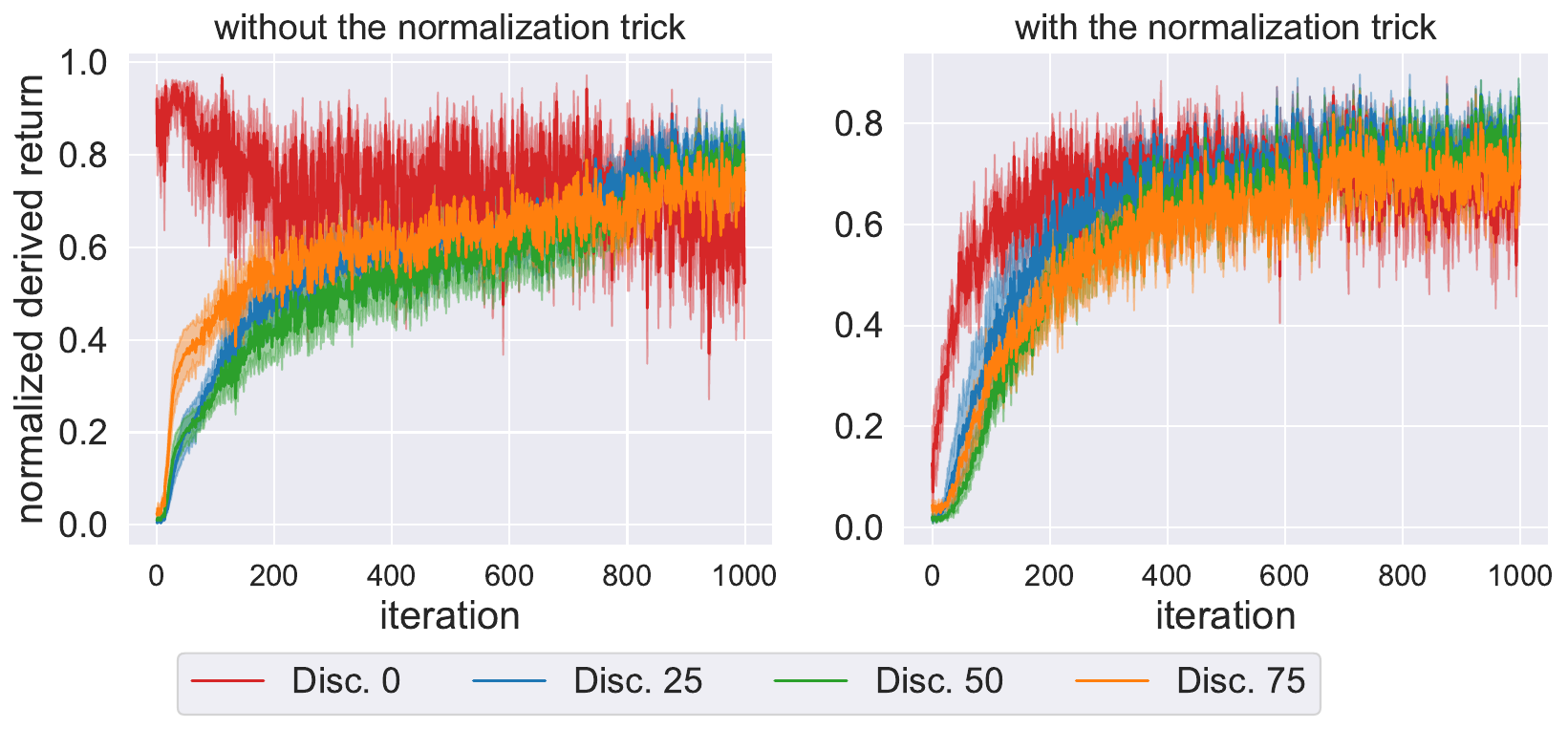}
    \caption{The return derived by the $\{0, 25, 50, 75\}$-th discriminator in \texttt{Ant} without (left) and with (right) the normalization trick. In both sub-figures, a policy is trained using the ensemble discriminator as the reward function, and the return derived by the $\{0, 25, 50, 75\}$-th discriminator is recorded. }
    \label{fig_changine_nature_influence}
\end{figure}

\subsection{Extended Dynamics-Agnostics Validation}

\label{supp_extened_dynamics_validata}

As a supplementary analysis to \cref{fig_dynamics_gravity_curve_fix_ret}, we expanded the range of \texttt{gravity} while keeping the true environment returns constant. To achieve this, we collected a large number of trajectories generated during policy training. From these, we selected those trajectories where the true environment return was approximately 75\% of the expert return, and plotted the results in \cref{fig_dynamics_gravity_curve_fix_ret_long}. The plot demonstrates that as the range of \texttt{gravity} increases and the policy diversity grows, DARL begins to vary with changes in \texttt{gravity}. However, compared to the reward model without the MI loss, DARL is significantly less affected by changes in dynamics. This indicates that the MI loss enables DARL to generate a reward model that is more decoupled from the dynamics.

\subsection{Memory and Time Cost of the Ensemble Model}
\label{supp_memory_cost}

\noindent{\textbf{Memory Cost.}} In \ours, we utilize an ensemble mechanism to construct the reward function. As we need to store and maintain historical discriminators, the increasing memory cost could pose a potential issue. However, in \texttt{Humanoid}, which has the largest state and action space dimensions among our tasks, the combined size of two encoder networks and a discriminator network is only 0.428 MB. Consequently, in all our experiments, the maximum memory cost for a single discriminator is $0.428$MB\footnote{Note that common models in computer vision, such as AlexNet and VGG11, have memory costs of $233.086$MB and $506.840$MB, respectively.}. We believe that the substantial performance improvement justifies this minor memory cost. 

While the memory cost of the discriminator ensemble was tolerable, it scales more rapidly compared to a single discriminator as model complexity increases. An alternative approach could be to replace the discriminator ensemble with last-iterate convergence no-regret learning methods. These methods do not require retaining historical discriminators and provide theoretical guarantees of convergence~\cite{golowich2020tight_last_iterate,lei2021last_iterate}. This substitution could be a valuable direction for future work.

\noindent{\textbf{Time Cost.}} In \ours, we infer policy rewards using an ensemble model, which can increase computation time linearly with the ensemble size. However, this computation can be parallelized using GPUs. We conducted an experiment to validate this approach, examining the relationship between the reward function computation time and GPU memory usage with varying numbers of discriminators. The results, shown in \cref{fig_time_cost}, used an RTX 4090 to parallelize the inference of multiple discriminators.

The results indicate that when the number of discriminators is fewer than 400, the reward inference time remains relatively constant, fluctuating around $1$ms. When the number of discriminators exceeds 400, the time overhead starts to rise gradually. Even with 1000 discriminators, the reward inference time is only around $2.5$ms, which is less than five times the inference time for a single discriminator. Thus, during the policy learning phase, there is no significant increase in training time as the number of discriminators grows.

Moreover, as shown in the right sub-figure of \cref{fig_time_cost}, the GPU memory overhead increases linearly with the number of discriminators. With 1000 discriminators, the total GPU memory overhead remains under $1$GB, well within the capacity of most GPUs.

\subsection{The Influence of the Changing Characteristics of Discriminators}
\label{app_influence_of_changing}
In this part, we analyze the impact of varying characteristics of the discriminator throughout the training process. Early-stage discriminators, which struggle to distinguish between expert and learner behaviors, typically output values around $0.5$. Conversely, discriminators that have reached convergence exhibit outputs close to $1.0$ for expert data and near $0.0$ for non-expert data. Thus, the output range of discriminators at later stages is significantly larger than at early stages. Consequently, the policy tends to disregard the output of early-stage discriminators. It is essential to recognize that early-stage discriminators are well-suited to lower-performance policies and can provide more accurate guidance for them. When the policy overlooks the output from early-stage discriminators, it might consider some actions inappropriate according to these early discriminators, but later-stage discriminators might mistakenly reward these actions. Consequently, the trained policy might occasionally execute suboptimal actions, affecting the overall performance.

Our ablation experiments on normalization techniques demonstrate that without balancing the output range disparity between different discriminators (\cref{fig_ablation_reward_transform}) results in unstable policy performance. To clarify this issue, we conducted two additional experiments. Following the ablation study results, we visualized the average scores given by discriminators to generated and expert data in each round within the \texttt{Ant} environment, as shown in \cref{fig_changing_chara}. Additionally, we also plotted the difference in scores between expert and generated data in \cref{fig_changing_chara}. Initially, the score difference between expert and generated data is minimal, nearly zero. However, as training progresses, this difference increases, with the disparity becoming significantly larger in later training stages.

To investigate the impact of output range disparity, we trained policies with and without the normalization technique and plotted the normalized return output by discriminators across training iterations, as shown in \cref{fig_changine_nature_influence}. Without the normalization technique, the output of the initial discriminator (Disc. 0) shows a decreasing trend, indicating that it has a minimal role in policy training and that the policy does not maximize its output. In contrast, when the output of different discriminators is balanced using the normalization technique, the policy gradually maximizes the score of the initial discriminator, demonstrating its equivalent importance to other models. This explains the observed performance drop when the normalization technique is not applied.

\end{document}